\documentclass{IEEEtran}

\usepackage{mathtools}
\usepackage{graphicx}
\usepackage{algorithm}
\usepackage{algpseudocode}
\usepackage{amsmath}
\usepackage{verbatim}
\usepackage{framed}
\usepackage{amssymb}

\usepackage{color}
\definecolor{linkBlue}{rgb}{.1,.1, .6}
\definecolor{linkGreen}{rgb}{.1,.35, .1}
\usepackage{hyperref}
\hypersetup{linkbordercolor = .7 .7 1 , 
                  citebordercolor = .5 1 .5,
                  colorlinks = true, 
                  linkcolor = linkBlue,  
                  citecolor = linkGreen ,
                  linktocpage=true,
                  pdftoolbar=true}

\newcommand{\labeledImage}[3]{
	\begin{minipage}{{#3}}
	\center {#2 \vspace{.05cm}}\\
	\includegraphics[width = {#3}]{{#1}}
	\end{minipage}~ 
}

\newcommand{\labeledImageBottom}[3]{
	\begin{minipage}{{#3}}
	\includegraphics[width = {#3}]{{#1}}
	\vspace{-.6cm}\center {#2}
	\end{minipage}~ 
}

\newcommand{\reals}{\mathbb{R}}

\newcommand{\half}{\frac{1}{2}}

\newcommand{\eqb}[1]{\begin{equation}\label{#1}}
\newcommand{\eqe}{\end{equation}}
\newcommand{\ra}{\rangle}
\newcommand{\la}{\langle}
\newcommand{\st}{\hbox{ subject to }}
\newcommand{\matb}{\left( \begin{matrix*}[r] }
\newcommand{\mate}{\end{matrix*}\right)}
\newcommand{\stone}{STOne }

\newcommand{\kp}{_{k+1}}

\DeclareMathOperator*{\argmax}{arg\,max}
\DeclareMathOperator*{\argmin}{arg\,min}
\newtheorem{theorem}{Theorem}
\newtheorem{lemma}{Lemma}

\newcommand{\sfn}[1]{S_{4^{#1}}}

\begin{document}

\bibliographystyle{ieeetr} 

\title{The STONE Transform:  Multi-Resolution Image Enhancement and Real-Time Compressive Video }
\author{Tom Goldstein, Lina Xu, Kevin F. Kelly, and Richard Baraniuk\\ }
\date{\today}
\maketitle

\begin{abstract}
Compressive sensing enables the reconstruction of high-resolution signals from under-sampled data.  While compressive methods simplify data acquisition, they require the solution of difficult recovery problems to make use of the resulting measurements.  This article presents a new sensing framework that combines the advantages of both conventional and compressive sensing.  Using the proposed \stone transform, measurements can be reconstructed instantly at Nyquist rates at any power-of-two resolution.  The same data can then be ``enhanced'' to higher resolutions using compressive methods that leverage sparsity to ``beat'' the Nyquist limit.  The availability of a fast direct reconstruction enables compressive measurements to be processed on small embedded devices.  We demonstrate this by constructing a real-time compressive video camera.   
\end{abstract}


\section{Introduction}

In light of the recent data deluge, compressive sensing has emerged as a method for gathering high-resolution data using dramatically fewer measurements.  This comes at a steep price.  While compressive methods simplify data acquisition, they require the solution of difficult recovery problems to make use of the resulting measurements. Compressive imaging has replaced the data deluge with an algorithmic avalanche --- conventional sensing saturates our ability to store information, and compressive sensing saturates our ability to recovery it.
   
   Spatial Multiplexing Cameras (SMC's) are an emerging technology allowing high-resolution images to be acquired using a single photo detector.  Interest in SMC's has been motivated by applications where sensor construction is extremely costly, such as imaging in the Short-Wave Infrared (SWIR) spectrum.  For such applications,  SPC's allow for the low-cost development of cameras with high resolution output.  
However the processing of compressive data is much more difficult, making real-time reconstruction intractable using current methods. 

  The burden of reconstruction is a major roadblock for compressive methods in applications.  For this reason it is common to reconstruct images offline when sufficient time and computing resources become available.  As a result, the need to extract real-time information from compressive data has led to methods for analyzing scenes in the compressive domain before reconstruction. Many of these methods work by applying image classifiers and other learning techniques directly on compressive data, sacrificing accuracy for computation tractability.

   The reconstruction problem is particularly crushing in the case of video processing. Naively extending SMC methods to video results in cameras with poor temporal resolution and burdensome Size, Weight and Power (SWaP) characteristics.   Many proposed video reconstruction schemes rely on costly \emph{optical flow} maps, which track the movement of objects through time.  This calculation makes video reconstruction slow ---  a few seconds of video may take hours or days to reconstruct.     When the number of frames becomes large, the reconstruction problem becomes orders of magnitude more costly than for still images, eclipsing the possibility of real-time video using conventional methods.

   We present a framework that unifies conventional and compressive imaging, capturing the advantages of both.   Using a new transform, we acquire compressive measurements that can be either reconstructed immediately at Nyquist rates, or ``enhanced'' using an offline compressive scheme that exploits sparsity to escape the Nyquist bound. 
   
\begin{figure}
\centering
Original\vspace{.1cm}\\
\begin{minipage}{8cm}
\includegraphics[width = 4cm]{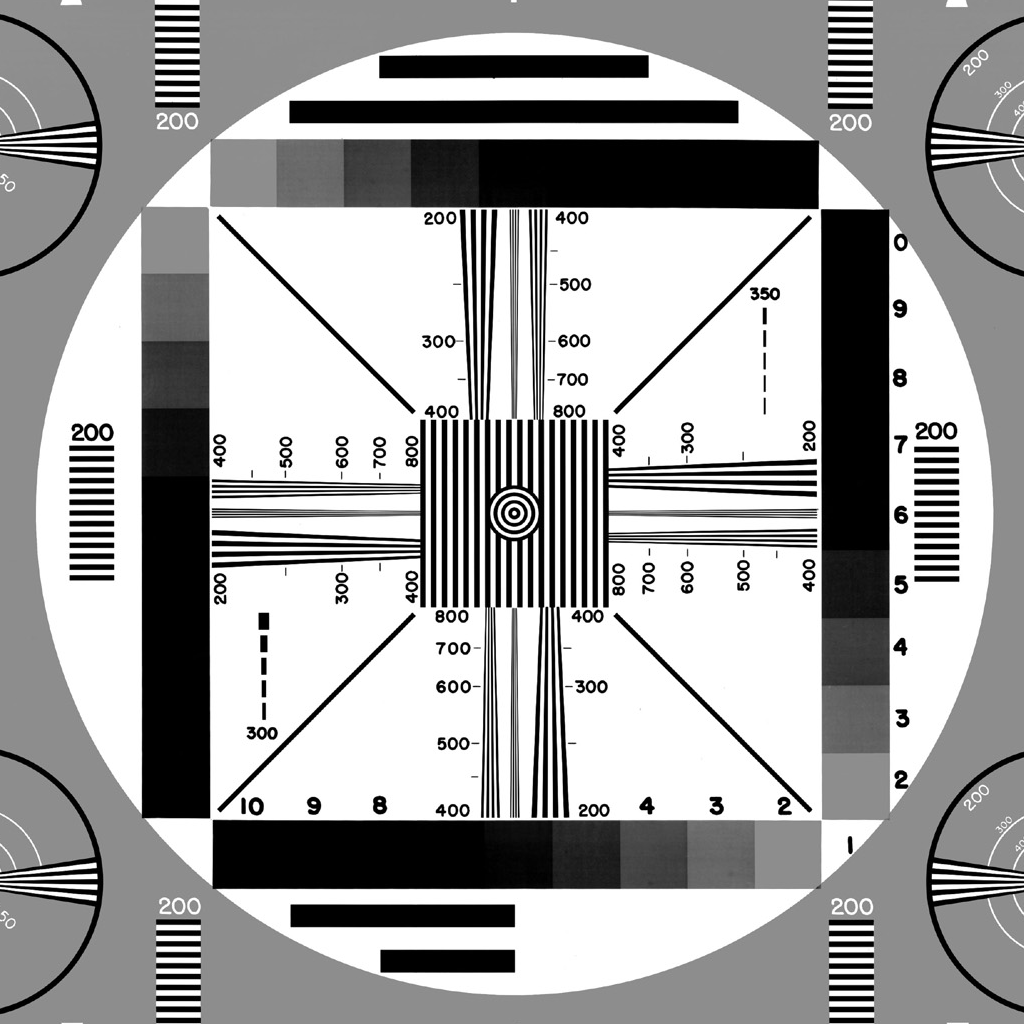}~
\includegraphics[trim=2cm 18cm 24cm 8cm,width = 4cm, clip]{testImage_original}
\end{minipage}\\
\center{Preview}\vspace{.1cm}\\
\begin{minipage}{8cm}
\includegraphics[width = 4cm]{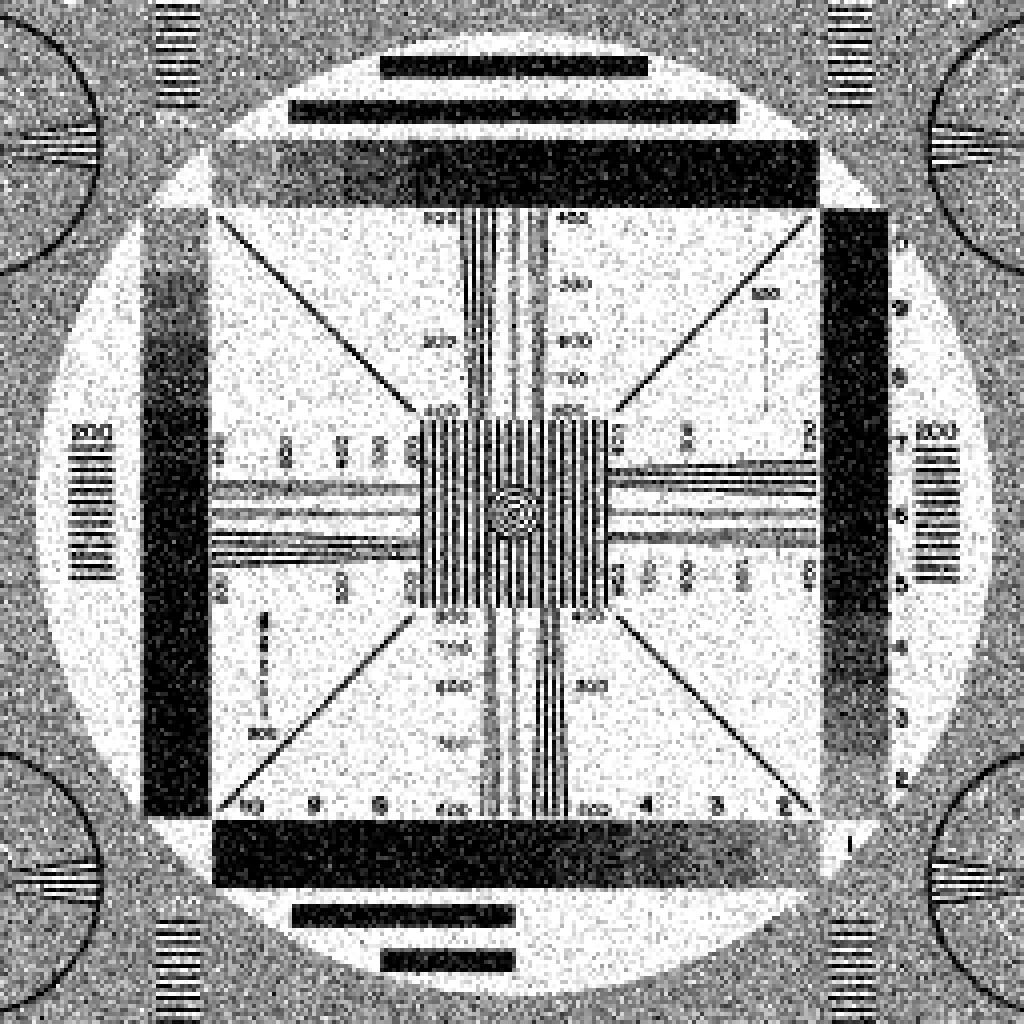}~
\includegraphics[trim=2cm 18cm 24cm 8cm,width = 4cm, clip]{testImage_preview}~
\end{minipage}\\
\center{Compressive}\vspace{.1cm}\\
\begin{minipage}{8cm}
\includegraphics[width = 4cm]{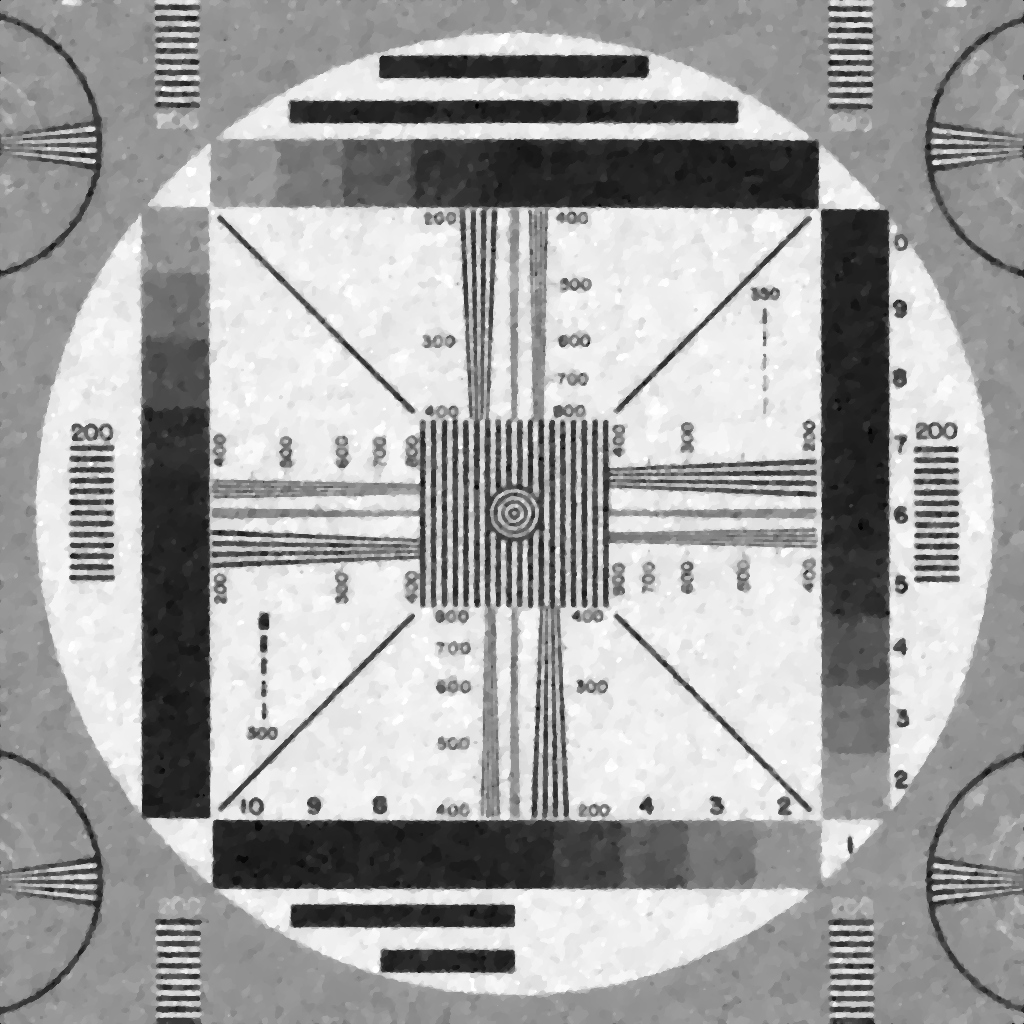}~
\includegraphics[trim=2cm 18cm 24cm 8cm,width = 4cm, clip]{testImage_compressive}
\end{minipage}
\caption{The proposed sensing operators allow under-sampled data to be reconstructed using either iterative/compressive or direct reconstructions.  (top) A 1024$\times$1024 test image.   (center) A direct (non-iterative) preview from $6.25\%$ sampling. (bottom)  An iterative reconstruction leverages compressed sensing to achieve higher resolution using the same data. }
\label{fig:example}
\end{figure}

   A cornerstone of our framework is a multi-scale sensing operator that enables reconstructed in two different ways.  First, the proposed measurements can be reconstructed using high-resolution compressive schemes that beat the Nyquist limit using sparse representations.  Alternately, the data can be reconstructed using a simple, direct fast  $O(N\log N)$ transform that produces ``preview'' images at standard Nyquist sampling rates.  This direct reconstruction transforms measurements into the image domain for online scene analysis and object classification in real time.  This way, image processing tasks can be performed without sacrificing accuracy by working in the compressive domain. The fast transform also produces real-time image and video previews with only trivial computational resources.  These two reconstruction methods are demonstrated on an under-sampled image in Figure \ref{fig:example}.

   In the case of high-resolution compressive video reconstruction, we also propose a numerical method that performs reconstruction using a sequence of efficient steps.  The compressive video reconstruction relies on a new ``3DTV'' model, which recovers video without expensive pre-processing steps such as optical flow.  In addition, the proposed reconstruction uses efficient primal-dual methods that do not require any expensive implicit sub-steps.   These numerical methods are very simple to implement and are suitable for real-time implementation using parallel architectures such a Field Programmable Gate Arrays (FPGA's).

  The flexibility of this reconstruction framework enables a real-time compressive video camera.  Using a single-pixel detector, this camera produces a data stream that is reconstructed in real time at Nyquist rates. After acquisition, the same video data can be enhanced to higher resolution using offline compressive reconstruction.

\subsection{Structure of this Paper}
 In Section 2, we present background information on compressive imaging and the challenges of compressive video.  In Section 3, we introduce the \stone transform, which enables images to be reconstructed using both compressive and Nyquist methods at multiple resolutions.  We analyze the statistical properties of the new sensing matrices in Section 4.  Then, we discuss the 3DTV model for compressive reconstruction in Section 5.  This model exploits compressive sensing to construct high resolution videos without common computational burdens.  Fast, simple numerical methods for compressive video reconstruction are introduced in Section 6, an assortment of applications are discussed in section 7, and numerical results are presented in Section 8.  

\section{Background}
  
\subsection{Single Pixel Cameras}
Numerous compressive imaging platforms have been proposed, including the Single-Pixel Camera \cite{DDTLSKB08}, flexible voxel camera \cite{GAVN10}, P2C2 \cite{RVC11}, and coded aperture arrays \cite{MHW09}.  To be concrete, we focus here on the spatially multiplexing Single Pixel Camera (SPC), as described in \cite{DDTLSKB08}.  However, the measurement operators and fast numerical schemes are easily applicable to a wide variety of cameras, including temporal and spectral multiplexing cameras.

Rather than measuring individual pixels alone, SPC's measure coded linear combinations of pixels \cite{DDTLSKB08}.  An SPC consists of a lens, a Digital Micro-mirror Device (DMD), and a photo detector.  Each mirror on the DMD modulates an individual pixel by diverting light either towards or away from the detector.  This results in a combination coefficient for that pixel of $+1$ or $-1,$ respectively.

 When the combination coefficients are chosen appropriately, the resulting measurements can be interpreted as transform coefficients (such as Hadamard coefficients) of the image.  For this reason, it is often said that SPC's sense images in the \emph{transform domain}.  

 The $i$th measurement of the device is an inner product $\la \phi_i,v \ra$ where $v$ is the vectorized image, and $\phi_i$ is a vector of $\pm1$'s encoding the orientation of the mirrors.  
Once $M$ measurements have been collected, the observed information can be written
$$\Phi v = b$$
where the rows of $\Phi$ contain the vectors $\{\phi_i\}$,  and $b$ is the vector of measurements.  If $v$ contains $N$ pixels, then $\Phi$ is an $M \times N$ matrix.  Image reconstruction requires of solving this system.  If $\Phi$ is a fast binary transform (such as a Hadamard transform) the image reconstruction is simple and fast.  However, it is often the case that $M<N,$ and image reconstruction requires additional assumptions on $v.$

\subsection{Compressed Sensing}
	Because SPC's acquire measurements in the transform domain (by measuring linear combinations of pixels), they can utilize \emph{Compressive Sensing} \cite{CRT06,Donoho06}, which exploits the compressibility of images to keep resolution high while sampling below the Nyquist rate.  Compressive imaging reduces the number of measurements needed for image reconstruction, and thus greatly accelerates imaging.   The cost of such compressive methods is that the image reconstruction processes is computationally intense.  Because we have fewer measurements than pixels, additional assumptions need to be made on the image.  

Compressed sensing assumes that images have sparse representations under some transform.  This assumption leads to the following reconstruction problem:
  \eqb{ell0}
  \min |S v|_0 \st \Phi v = b
  \eqe
where $S$ is the sparsifying transform, and the $\ell_0$ norm, $|\cdot|_0,$ simply counts the number of non-zero entries in $Sv.$  In plain words, we want to find the \emph{sparsest} image that is still compatible with our measurements.  It has been shown that when $S$ and $\Phi$ are chosen appropriately, nearly exact reconstructions of $v$ are possible using only $O( |S v|_0\log  N )$ measurements, which is substantially fewer than the $N$ required for conventional reconstruction.  In practice, accurate recovery is possible when $\Phi$ consists of randomly sampled rows of an orthogonal matrix \cite{CR06_2, BSN09}.

In practice, it is difficult to solve (\ref{ell0}) exactly.  Rather, we solve the convex relation, which is denoted:  
\eqb{energy} \min E(v) =  |S u|+  \frac{\mu}{2}\|\Phi u  - b \|^2 \eqe
for some regularization parameter $\mu,$ where $|\cdot|$/$\|\cdot\|$ denotes the $\ell_1$/$\ell_2$ norm, respectively.  It is known that when $S$ and $\Phi$ are \emph{mutually incoherent}, the solutions to  (\ref{energy})  and (\ref{ell0}) coincide with high probability.

Most often, the measurement matrix $\Phi$ is formed by subsampling the rows of an orthogonal matrix. In this case, we can write
$$\Phi = RT$$
where $R$ is a diagonal ``row selector'' matrix, with $R_{i,i}=1$ if row $i$ has been measured, and $R_{i,i}=0$ otherwise.  The matrix $T$ is generally an orthogonal matrix that can be computed quickly using a fast transform,  such as the Hadamard transform. 

\subsection{The Challenges of Compressive Video}

Much work has been done on compressive models for still-frame imaging, while relatively little is known about video reconstruction. 

   Video reconstruction poses many new challenges that still-frame imaging does not.  Object motion during data acquisition produces motion artifacts.  Rather than appearing as a simple motion blur (like in conventional pixel domain imaging), these motion artifacts get aliased during reconstruction and effect the entire image.   In order to prevent high levels of motion aliasing, reconstruction must occur at a high frame rate, yielding a small number of measurements per frame.
   
   The authors of \cite{Wakin12} conduct a thorough investigation of motion aliasing in compressive measurements.   A tradeoff exists between spatial resolution and temporal blur when objects are moving.  When images are sampled at low resolutions (i.e. pixels are large), objects appear to move slower relative to the pixel size, and thus aliasing due to motion is less severe.  When spatial resolution is high, motion aliasing is more severe.  For this reason it is desirable to have the flexibility to interpret data at multiple resolutions.  Higher resolution reconstructions can be used for slower moving and objects, and low resolutions can be used for fast moving objects.  This is an issues that will be addressed by the new Sum-To-One (\stone) transform, introduced in Section \ref{section:sto}.
  
\subsection{Previous Work on Compressive Video}
  A video can be viewed as a sequence of still frames, each individually sparse under some spatial transform.   However, adjacent video frames generally have high correlations between corresponding pixels.  Consequently a large amount of information can be obtained by exploiting these correlations.
  
One way to exploit correlations between adjacent frames is to use ``motion compensation.'' Park and Wakin \cite{PW13} first proposed the use of previews to compute motion fields between images, which could be used to enhance the results of compressive reconstruction.  

The first full-scale implementation on this concept is the Compressed Sensing Multi-scale Video (CS-MUVI) framework \cite{SSB12}.  This method reconstructs video in three stages.  First, low-resolution previews are constructed for each frame.  Next, an optical flow model is used to match corresponding pixels in adjacent frames. Finally, a reconstruction of the type (\ref{blocks},\ref{energy}) is used with ``optical flow constraints'' added to enforce equality of corresponding pixels.

The CS-MUVI framework produces good video quality at high compression rates, however it suffers from extreme computational costs. The CS-MUVI framework relies on special measurement operators that cannot be computed using fast transforms. A full-scale transform requires $O(N^2)$ operators, although the authors of \cite{SSB12} use a ``partial'' transform to partially mitigate this complexity.  In addition, optical flow calculation is very expensive, and is a necessary ingredients to obtain precise knowledge of the correspondence between pixels in adjacent images.   Finally, optimization schemes that can handle relatively unstructured optical flow constraints are less efficient than solvers for more structured problems.  

A similar approach was adopted by the authors of \cite{AHBR13} in the context of Magnetic Resonance Imaging (MRI). Rather than use previews to generate motion maps, the authors propose an iterative process that alternates between computing compressive reconstructions, and updating the estimated flow maps between adjacent frames.  This method does not rely on special sensing operators to generate previews, and works well using the Fourier transform operators needed for MRI. However, this flexibility comes at a high computational cost, as the computation of flow maps must now be done iteratively (as opposed to just once using low-resolution previews). Also, the resulting flow constraints have no regular structure that can be exploited to speed up the numerics.

For these reasons, it is desirable to have more efficient sparsity models allowing for efficient computation.  Such a model should (1) not require a separate pre-processing stage to build the sparsity model, (2) rely on simple, efficient numerical schemes, and (3) use only fast transform operators that can be computed quickly.  The desire for more efficient reconstruction motivates the 3DTV framework proposed below.  

\section{Multi-Resolution Measurement Matrices: The \stone Transform} \label{section:sto}
In this section, we discuss the construction of measurement operators that allow for fast low-resolution ``previews.''  Previews are conventional (e.g. non-compressive)  reconstructions that require one measurement per pixel.  Because they rely on direct reconstructions, previews have the advantage that they do not require the solution of expensive numerical problems and thus require less power and time to reconstruct.  At the same time, the data used to generate previews is appropriate for high-resolution compressive reconstruction when the required resources are available.  

Such measurement matrices have the special property that each row of the matrix has unit sum, and so we call this the Sum To One (STOne) transform. 
  
\subsection{The Image Preview Equation}

Previews are direct reconstructions of an image without the need for full-scale iterative methods.  
Such an efficient reconstruction is only possible if the sensing matrix $\Phi$ is designed appropriately.  We wish to identify the conditions on $\Phi$ that make previews possible.  
 
The low resolution preview must be constructed in such a way that it is compatible with measured data. In order to measure the quality of our preview using high resolution data, we must convert the low-resolution $n\times n$ image into a high resolution $N\times N$ image using the prolongation operator $P_n^N$.  To be compatible with our measurements, the low-resolution preview must satisfy
\eqb{previewEquation}
\Phi P u_n = b
\eqe
where we have left the sub/super-scripts off $P$ for notational simplicity.  In words, when the preview is up-sampled to full resolution, the result must be compatible with the measurements we have taken.

The solution of the preview equation (\ref{previewEquation}) is highly non-trivial if the sensing matrix is not carefully constructed.  Even when $\Phi$ is well-behaved, it may be the case that $\Phi P$ is not.  For example, if $\Phi$ is constructed from randomly selected rows of a Hadamard matrix, the matrix  $\Phi P$ is poorly-conditioned.  As a result, the low-resolution preview is extremely noise-sensitive and inaccurate.  Also, for many conventional sensing operators, there is no efficient algorithm for solving the preview equation.  Even when $\Phi$ can be evaluated efficiently (using, e.g., the fast Hadamard or Fourier transform), solving the preview equations may require slow iterative methods.

Ideally, we would like the matrix $\Phi P$ to be unitary.  In this case, the system is well-conditioned making the preview highly robust to noise.  Also,  the equations can be explicitly solved in the form $u_n = \Phi^TP^Tb.$

These observations motivate the following list of properties that a good sensing operator should have:

\begin{enumerate}
\item \emph{The matrix $\Phi$ has good compressive sensing properties.}  It is known that compressive reconstruction is possible if the rows of $\Phi$ are sub-sampled from an orthogonal matrix \cite{CR06_2, BSN09}.  That is, we want to define $\Phi = RT$ for some row selector $R$ and fast orthogonal transform $T$.
\item  \emph{The preview matrix $\Phi P$ must be well conditioned and easily invertible.} These conditions are ensured if $\Phi P$ is unitary.  
\item  \emph{The entries in $\Phi$ must be $\pm1$}. Our sensing matrix must be realizable using a SMC.
\end{enumerate}

 It is not clear that a sensing matrix possessing all these properties exists, and for this reason several authors have proposed sensing methods that sacrifice at least one of the above properties.  CS-MUVI, for example, relies on Dual-Scale-Space (DSS) matrices that satisfy properties 2 and 3, but not 1.  For this reason, the DSS matrices do not have  a fast transform, making reconstruction slow.
 
Below, we describe the construction of sensing matrices that satisfy all of the above desired properties.

\subsection{Embeddings of Images into Vectors}
Suppose we have compressive measurements taken from an $N \times N$ image.  We wish to acquire a low-resolution $n \times n$ preview for $n<N.$  If $n$ evenly divides $N,$ then we can define the downsampling ratio $\delta = N/n.$

Depending on the situation, we will need to represent images as either a 2-dimensional array of pixels, or as a 1-dimensional column vector.     The most obvious embedding of images into vectors is using the row/column major ordering, or equivalently to perform the transform on the image in the row and column directions separately.  This embedding does not allow low-resolution previews to be constructed using a simple transform.

 Rather, we embed the image into a vector by evenly dividing the image into blocks of size $\delta\times \delta.$  There will be $n^2$ such blocks.  The image is then vectorized block-by-block. The resulting vector has the form
\eqb{vec}
v =\left( \begin{matrix}
v_1\\ v_2\\\vdots\\v_{n^2}
\end{matrix}\right)
\eqe
	where $v_i$ contains the pixel data from the $i$th block.

It is possible to embed the image so that the vector is in block form (\ref{vec}) for  every choice of $n=2^k<N$. 
  When such an embedding is used, previews can be obtained at arbitrary resolutions.  

  The new embedding is closely related to the so-called \emph{nested-dissection} ordering of the image, which is well known in the numerical linear algebra literature \cite{saad03,George73}.   The proposed ordering is defined by a recursive function which breaks the square image into four symmetrical panels.  Each of the four panels are addressed one at a time.    Every pixel in the first panel is numbered, and then the second panel, and then the third and fourth.  The numbering assigned to the pixels in each panel is defined by applying the recursive algorithm.  Pseudocode for this method is given in Algorithm \ref{alg:pixOrder}.

\begin{figure*}
\label{fig:pixOrder}
\caption{The recursive algorithm for embedding a $16\times 16$ image into a 1-dimensional signal.  The ordering of the pixels in generated by recursively breaking the image into blocks and numbering each block individually.  (Top Left) The first stage of the algorithm breaks the image into 4 blocks, and assigns a block of indices to each in clockwise order. (Bottom Left)  The second level of recursion breaks each block into 4 sub-panels, which are each numbered in clockwise order.  (Right)  The third level of recursion breaks each sub-panel into 4 pixels, which are numbered in clockwise order.  Note that the ordering of the panels is arbitrary, but we choose the clockwise ordering here for clarity.}
\includegraphics[width = \linewidth]{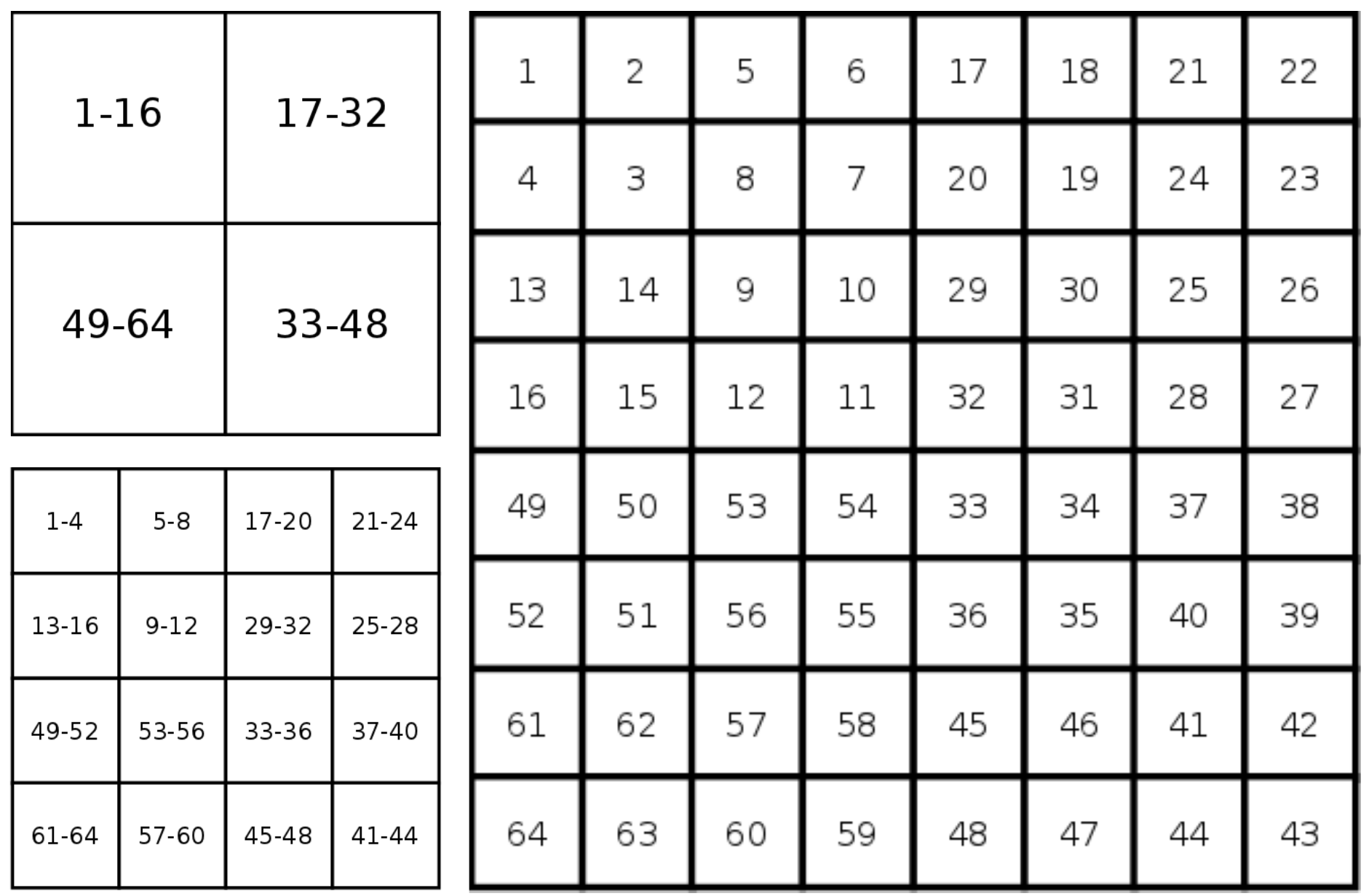}
\end{figure*}

 \begin{algorithm}
    \caption{Ordering of Pixels in an $n\times n$ Image}
    \label{alg:pixOrder}
    \begin{algorithmic}[1]
    \State Inputs: 
    \State \hspace{.5cm} \textbf{Image}: a 2d array of pixels
    \State \hspace{.5cm}  \textbf{L}:    The lowest index to be assigned to a pixel
   	\Function{assign}{\textbf{Image},\textbf{L}}
   	\State Let $N$ be the side length of \textbf{Image}
   	\If{$N=1$}
   	\State \textbf{Image} = L
   	\State \Return
   	\EndIf
   	\State Break \textbf{Image} into four panels, \{$\mathbf{I}_1,$ $\mathbf{I}_2, $ $\mathbf{I}_3,$ $\mathbf{I}_4$\}
   	\State \Call{assign}{$\mathbf{I}_1,\mathbf{L}$}
   	\State \Call{assign}{$\mathbf{I}_2,\mathbf{L}+N\times N/4$}
    \State \Call{assign}{$\mathbf{I}_3,\mathbf{L}+2N\times N/4$}
   	\State \Call{assign}{$\mathbf{I}_4,\mathbf{L}+3N\times N/4$}
    \EndFunction
       \end{algorithmic}
  \end{algorithm}

This recursive process is depicted graphically in figure (\ref{fig:pixOrder}).   Note that at every level of the recursion, we have assigned a contiguous block of indices to each sub-block of the image, and thus the corresponding embedding admits a decomposition of the form  (\ref{vec}).

Below, we present Theorem (\ref{thrm:recon}), which can be used to obtain previews for any $n = 2^k$ with $1\le k<K.$  


\subsection{Interpolation Operators}
	In order to study the relationship between the low and high resolution images, we will need prolongation/interpolation operators to convert between these two resolutions.

  The prolongation operator maps a small $n\times n$ image into a large $N \times N$ image.  It replaces each pixel in the low-resolution image with a $\delta \times \delta$ block of pixels.  If the image has been vectorized in the block-wise fashion described above, the prolongation operator can be written in block form as
  $$P_n^N = 
  \left( \begin{matrix}
  1_{\delta^2}	 & 0 & \cdots  & 0\\
  0  & 1_{\delta^2}	&   \cdots  & 0\\
   \vdots&  & \ddots & \vdots \\
 0  & 0 &  \cdots  & 1_{\delta^2}
   \end{matrix}\right) = I_{ n^2 }\otimes 1_{\delta^2}$$
   where $1_{\delta^2}$ denotes a $\delta^2\times 1$ column vector of 1's, and $\otimes$ denotes the Kronecker product.  
   
   The row-selector matrix and measurement vector can also be broken down into blocks.  We can write
     $$R = \matb R_1 & 0& \cdots&0 \\ 
     					0 & R_1& \cdots&0 \\ 
     					 \vdots & & \ddots & \vdots\\
						0 & 0& \cdots&R_{n\times n} \\ 
     				\mate, \qquad
     	b =\left( \begin{matrix}
						b_1\\ b_2\\\vdots\\b_{n^2}
					\end{matrix}\right)
	$$
     				where $R_i$ is a $\delta^2\times \delta^2$ row-selector sub-matrix, and $b_i$ contains a block of $\delta^2$ measurements.

\subsection{A Sum-To-One Transform}

We now introduce a fast orthogonal transform that will be the building block of our sensing matrices. This transform has the property that each row of the transform matrix Sums To One, and thus we call it the \stone transform.  It will be shown later that this property is essential for the existence of fast preview reconstruction methods.

Consider the following matrix stencil:

$$S_4 = \half \matb
-1 & 1 &1 &1  \\ 
 1 & -1 &1 &1  \\ 
 1 & 1 &-1 &1  \\
 1 & 1 &1 &-1  \\  
\mate$$
It is clear by simple observation that this matrix is unitary (i.e. $S_4^TS_4 = I_4$), and its Eigenvalues are $\pm1. $  Furthermore, unlike the stencil for the standard Hadamard matrix, this stencil has the property that the rows of the matrix sum to 1.

We will use the matrix $S_4$ as a stencil to construct a new set of transform matrices as follows
  $$\sfn{k+1} = S_4  \otimes \sfn{k} =  \half \matb
-\sfn{k} & \sfn{k} &\sfn{k} &\sfn{k}  \\ 
 \sfn{k} & -\sfn{k} &\sfn{k} &\sfn{k}  \\ 
 \sfn{k} & \sfn{k} &-\sfn{k} &\sfn{k}  \\
 \sfn{k} & \sfn{k} &\sfn{k} &-\sfn{k}  \\  
\mate$$
  where $\otimes$ denotes the Kronecker product.
  We have the following result;
  \begin{theorem}
  	For all $k \ge1,$ each row of the matrix  $S_{4^k}$ sums to 1.  Also, every matrix $S_{4^k}$ is unitary.
  \end{theorem}
\begin{IEEEproof}
  The summation result follows immediately by inspection.  To prove the unitary claim, we simply use the Kronecker product identity  $(A \otimes B) (C \otimes D) = AC \otimes BD$.  It follows that
  
    \begin{eqnarray*}\sfn{k+1}^T\sfn{k+1} &=& (S_4\otimes \sfn{k})^T(S_4\otimes \sfn{k}) \\&=& (S_4^T\otimes \sfn{k}^T)(S_4\otimes \sfn{k}) = S_4^TS_4 \otimes \sfn{k}^T \sfn{k}\\
    &=&I_4 \otimes I_{4^k}=I_{4^{k+1}}.  
 	\end{eqnarray*}
 	The result follows by induction.
  \end{IEEEproof}
Note that because the Kronecker product is associative,  we can form arbitrary decompositions of $\sfn{k}$ of the form 
$$\sfn{k} = \sfn{k-\ell}\otimes \sfn{\ell}$$
for any $\ell$ between $1$ and $k$.

\subsection{Reconstructing Low Resolution Previews}  \label{section:recon}
In this section, we show how the sum-to-one transform can be used to obtain low-resolution previews.  The construction works at any power-of-two resolution.

   The low resolution preview is possible if our data satisfies the following simple properties:
   \begin{enumerate}
   \item  The sensing matrix is of the form $$\Phi = RS_{N^2},$$ 
   where $R$ is a row-selector matrix, and $S_{N^2}$ is the fast sum-to-one transform.
   \item Every block, $R_i,$ of $\delta^2$ diagonal entries of $R$ contains at least one non-zero entry.
   \item Every $\delta\times \delta$ patch of the image is mapped contiguously into the vector $v$.  The will be true for any power-of-two downsampling when the pixels are ordered using Algorithm \ref{alg:pixOrder}.
   \end{enumerate}

 We show below that when the measurement operator has these two properties it is possible to efficiently recover low-resolution previews using a simple fast transform.  
   
   As discussed above, the low-resolution preview requires us to solve the equation 
\eqb{previewEquation2}
\Phi P u_n = R S_{N^2}P u_n = b
\eqe
When the measurements are taken using the sum-to-one transform, the solution to this equation is given explicitly by the following theorem.

\begin{theorem} \label{thrm:recon}
	Consider a row selector, $R,$ a measurement operator of the form $\Phi = RS_{N^2},$ and a prolongation operator P as described above.   Suppose that images are represented as vectors in the block form (\ref{vec}).
	
	 If each sub-matrix $R_i$ contains exactly one non-zero entry, then the preview equation has a unique solution, which is given by 
	   $$u_n = S_{n^2} \bar{b},$$  
	   where $\bar b$ is an $n^2\times 1$ vector containing the known entries of $b.$
	   
	   If each sub-matrix $R_i$ contains one or more non-zero entries, then the preview equation may be overdetermined, and the least squares solution is given by
	     $$u_n = S_{n^2} \hat{b},$$  
	     where $\hat b_i = \left (\sum_{\gamma=1}^{\delta^2}b_{i,\gamma}\right) / \left( \sum_{\gamma=1}^{\delta^2} R_i \right).$  In other words, $\hat b_i$ is the mean value of the known entries in $b_i.$     
\end{theorem}
\begin{IEEEproof}
The preview equation contains the product $S_{N^2} P,$ which we can decompose using the Kronecker product definition of the sum-to-one transform:
$$ S_{N^2} P = (S_{n^2}\otimes S_{N^2/n^2})P = (S_{n^2}\otimes S_{\delta^2}) P.$$

The up-sampling operator works by replacing every pixel in $u_n$ with a constant panel of size $\delta \times \delta,$ and has the form  $P =  I_{ n^2 }\otimes 1_{\delta^2}.$
We can thus write
\begin{align*}  S_{N^2} P   &= (S_{n^2}\otimes S_{\delta^2}) (I_{n^2 }\otimes 1_{\delta^2}) \\
 &= (S_{n^2}     I_{n^2 } )\otimes ( S_{\delta^2} 1_{\delta^2}) \\
 & = S_{n^2} \otimes ( S_{\delta^2} 1_{\delta^2}).
\end{align*}
Now is when the sum-to-one property comes into play.  The product $S_{\delta^2} 1_{\delta^2}$ simply computes the row sums of the matrix $S_{\delta^2},$ and so  $S_{\delta^2} 1_{\delta^2} = 1_{\delta^2}.$     This gives us 
$$  S_{N^2} P   = S_{n^2} \otimes  1_{\delta^2}.$$

Using this reduction of the transform and prolongation operators, the low-resolution preview equation (\ref{previewEquation2}) reduces to
$$\Phi P u_n = R S_{N^2}P u_n = R (S_{n^2} \otimes  1_{\delta^2}) u_n.$$
Note that the matrix  $(S_{n^2} \otimes  1_{\delta^2})$  is formed by taking the matrix $S_{n^2}$ and copying each of its rows $\delta^2$ times.  If each of the $R_i$ contain only a single non-zero entry, then the operator $R$ simply selects out a single copy of each row.  In this case, it follows that the preview equation reduces to
$$\Phi P u_n = S_{n^2} u_n = \bar b.$$
The solution to this equation is $u_n =S_{n^2}^{-1}  \bar b =S_{n^2}  \bar b.$

If we allow the $R_i$ to contain more than one non-zero element, then we must seek the least squares solution.  In this case, consider the sum-of-squares error function:
  \begin{align}\begin{split}\label{ls}
  E(u) &= \| R (S_{n^2} \otimes  1_{\delta^2}) u - b\|^2 \\
    &= \sum_{i=1}^{n^2}\sum_{\gamma=1}^{\delta^2}  R_{i,\gamma}(S_{n^2,i}\cdot u - b_{i,\gamma})^2 
  \end{split}\end{align}
   where $S_{n^2,i}$ denotes the $i$th row of $S_{n^2}$.
Observe now that 
\begin{samepage}
\begin{align*}
\sum_{\gamma=1}^{\delta^2} R_{i,\gamma} (S_{n^2,i}\cdot u - b_{i,\gamma})^2 
     &= \sum_{\gamma=1}^{\delta^2} R_{i,\gamma}\| S_{n^2,i}\cdot u\|^2\\
     &- 2R_{i,\gamma}\la S_{n^2,i},b_{i,\gamma} \ra+R_{i,\gamma}\|b_{i,\gamma}\|^2
     \end{align*}
      \begin{eqnarray*}
       &=&r_i \| S_{n^2,i}\cdot u\|^2- 2\la S_{n^2,i}, \sum_{\gamma=1}^{\delta^2}b_{i,\gamma} \ra+ \sum_{\gamma=1}^{\delta^2}\|b_{i,\gamma}\|^2\\
        &=&\left (\sqrt{r_i}  S_{n^2,i}\cdot u -  \frac{1}{\sqrt{r_i}}\sum_{\gamma=1}^{\delta^2}b_{i,\gamma} \right)^2\\
                 &&+ \sum_{\gamma=1}^{\delta^2}\|b_{i,\gamma}\|^2 - \frac{1}{r_i} \left \|\sum_{\gamma=1}^{\delta^2}b_{i,\gamma} \right\|^2\\
        &=& \left ( \sqrt{r_i}  S_{n^2,i}\cdot u -  \frac{1}{\sqrt{r_i}}\sum_{\gamma=1}^{\delta^2}b       _{i,\gamma} \right)^2+C_i
      \end{eqnarray*} 
      \end{samepage}

\noindent where  $C_i$ is a constant that depends only on $b$.  It follows that the least-squares energy (\ref{ls}) can be written

\begin{align*}
  E(u) &=  \sum_{i=1}^{n^2} (\sqrt{ r_i}  S_{n^2,i}\cdot u -  \frac{1}{\sqrt{r_i}}\sum_{\gamma=1}^{\delta^2}b_{i,\gamma} )^2+ \sum_{i=1}^{n^2}C_i \\
  &= \| \hat R^{1/2} S_{n^2} u - \hat R^{1/2} \hat b \|^2+C. 
  \end{align*}
where $C$ is a constant, $\hat R_i = r_i,$ and $\hat b_i = \frac{1}{r_i}\sum_{\gamma=1}^{\delta^2}b_{i,\gamma}.$
  This energy is minimized when we choose $u_n$ to satisfy the equation
   $$S_{n^2} u = \hat b.$$
\end{IEEEproof}

\begin{figure*}
  \includegraphics[width=\linewidth, trim = 15 90 100 0, clip]{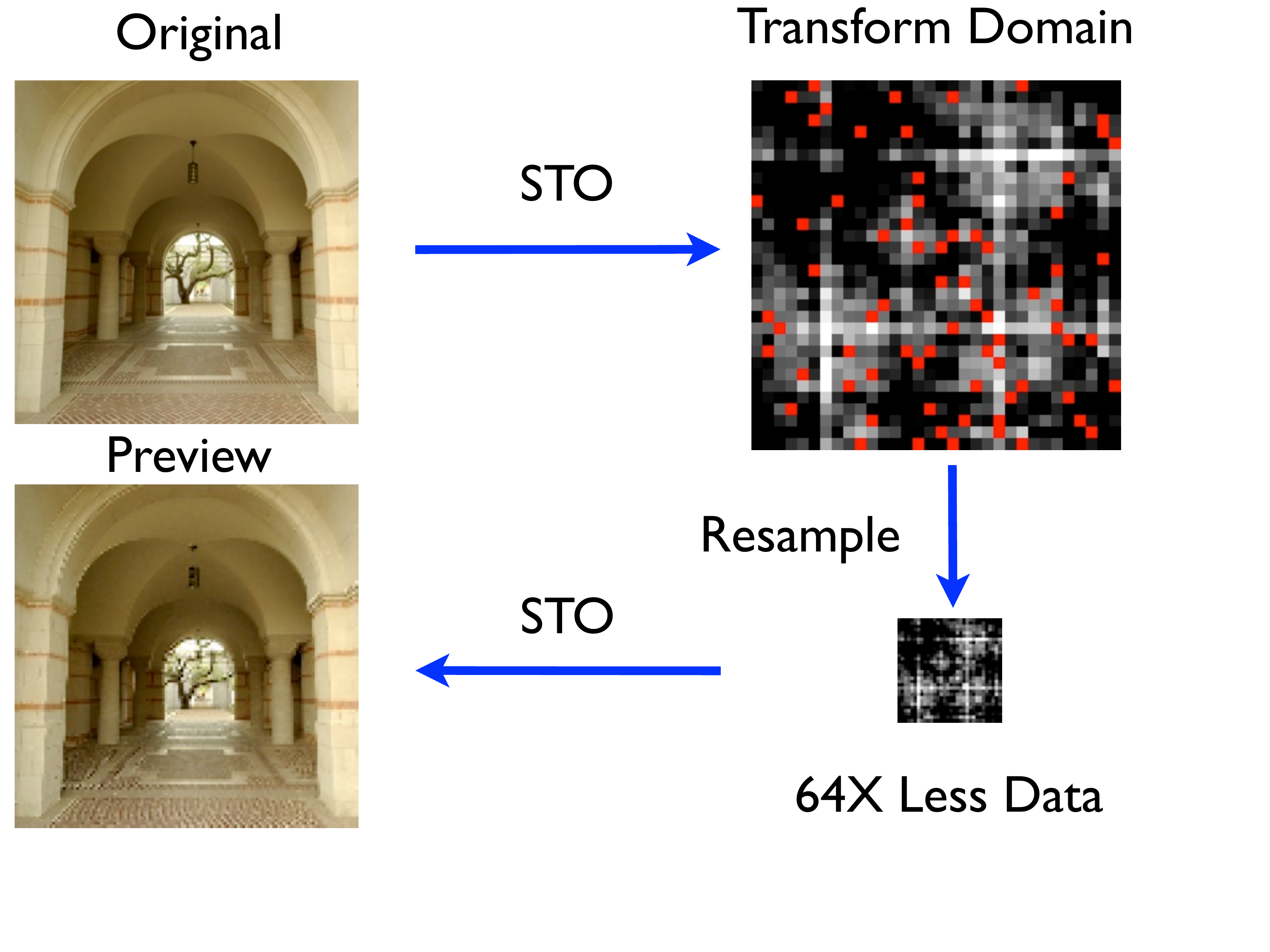}
  \caption{The reconstruction of the low resolution preview.  The original image is measured in the transform domain where the \stone coefficients are sub-sampled.  The sub-sampled coefficients are then re-binned into a low resolution \stone domain, $\bar b.$  Finally, the low-resolution transform is inverted to obtain the preview. }
   \label{fig:preview}
  \end{figure*}

Theorem \ref{thrm:recon} shows that under-sampled high resolution \stone coefficients can be re-binned into a complete set of low resolution \stone transform coefficients.  These low-resolution coefficients are then converted into an image using a single fast transform.    This process is depicted in Figure \ref{fig:preview}. 

\subsection{Design of the Measurement Sequence}
  In this section, we propose a ``structured random'' measurement sequence that allows for previews to be obtained at any resolution or time. 
  It is clear from the discussion above that it is desirable to form the measurement operator by subsampling rows of $S_N.$  However, the order in which these rows are sampled is of practical significance, and must be carefully chosen if we are to reconstruct previews at arbitrary resolutions.   We demonstrate this by considering the following scenarios.
    
    Suppose that we have a sequence of measurements $\{b_{k_i}\}$ taken using row indices $K=\{k_i\}$ of the matrix $S_{N^2}$.  After measurement $b_i$ is acquired, we wish to obtain an $n\times n$ preview using only the most recent $n^2$ measurements.  If we break the set of rows of $S_N^2$ into $n^2$ groups, then we know from Theorem \ref{thrm:recon} that the preview exists only if we have sampled a measurement from each group.  It follows that each of latest $n^2$  measurements must lie in a unique group.    
   
     After measurement $b_{i+1}$ is acquired, the measurement window is shifted forward.  Since we want to have previews available at any time, it must still be true that the most recent $n^2$ measurements lie in unique groups even after this shift.  
     
     Now, suppose that because of fast moving objects in the image, we decide we want a ``faster'' preview with a shorter acquisition window.  We now need an $n' \times n'$ preview with $n'<n$.  When we sample the most recent $(n')^2$ data and redistribute the row space into $(n')^2$ groups, each of the data in this new window must lie in a unique group.
 
  Clearly, the measurement sequence must be carefully constructed to allow for this high level of flexibility.  The measurement sequence must have the property that, for any index $i$ and resolution $n$, if we break the row space into $n^2$ groups, the measurements $b_{i-n^2+1}$ through $b_i$ all lie in separate groups.  Such an ordering is produced by the recursive process listed in Algorithm \ref{alg:modeOrder}. 
    
 \begin{algorithm}
    \caption{Sampling Order of Rows From \stone Matrix}
    \label{alg:modeOrder}
    \begin{algorithmic}[1]
    \State Inputs: 
    \State \hspace{.5cm} $K = \{K_i\}$: a linearly ordered list of row indices
   	\Function{order}{K}
   	\State Let $|K|$ be the length of the list $K$.
   	\If{$|K|=1$}
   	\State \Return K
   	\EndIf
   	\State Form 4 sub-lists: $ \{k_i\}_{i=1}^{|K|/4},$ $\{k_i\}_{i=|K|/4+1}^{|K|/2},$ $\{k_i\}_{i=|K|/2+1}^{3|K|/4},$ $\{k_i\}_{i=3|K|/4+1}^{|K|}$
   	\State Randomly assign the names $K^1,$ $K^2,$ $K^3,$ and $K^4$ to sub-lists
   	\State \Call{order}{$K^1$}
   	\State \Call{order}{$K^2$}
  	\State \Call{order}{$K^3$}
   	\State \Call{order}{$K^4$}
   	\State $K \leftarrow \{K^1_1 K^2_1 K^3_1 K^4_1K^1_2 K^2_2 K^3_2 K^4_2\cdots  K^4_{\frac{|K|}{4}} \} $
    \EndFunction
       \end{algorithmic}
  \end{algorithm}
  
      The top level input to the algorithm is a linearly ordered sequence of row indices, $K = \{k_i\}=\{1,2,\cdots N^2\}.$  At each level of the recursion, the list of indices is broken evenly into four parts,   $ \{k_i\}_{i=1}^{|K|/4},$ $\{k_i\}_{i=|K|/4+1}^{|K|/2},$ $\{k_i\}_{i=|K|/2+1}^{3|K|/4},$ and $\{k_i\}_{i=3|K|/4+1}^{|K|},$ and each group is randomly assigned a unique name from $\{K^1,K^2,K^3,K^4\}.$  Each of the 4 groups is then reordered by recursively calling the reordering algorithm.  Once each group has been reordered, they are recombined into a single sequence by interleaving the indices - i.e. the new sequence contains the first entry of each list, followed by the second entry of each list, etc.  
 
 Note that Algorithm \ref{alg:modeOrder} is non-deterministic because the index groups formed at each level are randomly permuted.  Because of these random permutations, the resulting ordering $K$  exhibits a high degree of randomness suitable for compressive sensing.
 
   Compressive data acquisition proceeds by obtaining a sequence of measurements $b_i = \la S_{N^2,k_i} v \ra,$ where $S_{N^2,k_i} $ denotes row $k_i$ of $S_{N^2},$ and the sequence $K=\{k_i\}$ is generated from Algorithm \ref{alg:modeOrder}.  If the number of data acquired exceeds $N^2,$ then we proceed by starting over with row $k_0,$ and proceeding as before.  Any window of length $n^2$ will still contain exactly one element from each row group, even if the window contains the measurement where we looped back to $k_0.$

\section{Statistical Analysis of the \stone Preview}

Suppose we break an $N\times N$ image into an $n\times n$ array of patches.  In the corresponding low resolution preview, each pixel ``represents''  its corresponding patch in the full-resolution image.  The question now arises:  How accurately does the low resolution preview represent the high resolution image?  
   
    We can answer this question by looking at the statistics of the low resolution previews.  When the row selector matrix is generated at random, the pixels in the preview can be interpreted as random variables.  In this case, we can show that the expected value of each pixel equals the mean of the patch that it represents.  Furthermore, the variation of each pixel about the mean is no worse than if we had chosen a pixel at random as a representative from each patch.  To prove this, we need the following simple lemma:
      
  \begin{lemma} \label{lemma:stat}
 The $i$th entry of $S_N v$ can be written
 $$S_N(i)v = \mu + e(i)$$
 where $\mu = 1_NMean(v),$  $Mean(e) =0,$ and  $Var(e) = Var(v).$
\end{lemma}
\begin{IEEEproof} 
We have $S_N(i)v  = \sum_j S_N(i,j) v(j),$
 and so 
 \begin{align*}
 Mean(S_Nv) &= \frac{1}{N}\sum_i S_N(i)v =  \frac{1}{N}\sum_i   \sum_j S_N(i,j) v(j) \\
   &=   \frac{1}{N}\sum_j v(j)  \sum_i S_N(i,j) =  \frac{1}{N}\sum_j v(j) = \mu.  
 \end{align*}
It follows that $Mean(e) = Mean(S_Nv - \mu)=0.$  

  The variation about the mean is given by $$Var(e) = Var(S_Nv-\mu) = Var(v-\mu)=Var(v)$$
  where we have used the unitary nature of $S_N$ and sum to one property, which gives us the identity $S_N\mu = \mu$. 
  \end{IEEEproof}
  
  We now prove the statistical accuracy of the preview.
\begin{theorem} \label{thrm:preview}
Suppose we create an $n\times n$ preview from an $N\times N$ image.  Divide the rows of $S_{N^2}$ into $n^2$ groups.  Generate the row selector matrix by choosing an entry uniformly at random from each group.  Then the expected value of each pixel in the low resolution preview equals the mean of the corresponding image patch.  The variance of each pixel about this mean equals the mean of the patch variances .
\end{theorem}
\begin{IEEEproof} 
 The \stone transform has the following decomposition:
$$\sfn{K} = \sfn{K-k}\otimes \sfn{k}.$$
Let $\delta = N/n.$   Break the rows of $\sfn{K}$ into $n^2$ groups, each of length $\delta^2.$ The $r$th row block of $\sfn{K}$ can then be written (using ``Matlab notation'')
\begin{align*}
\sfn{K}
&(r\delta^2:r\delta^2+\delta^2,:)v 
 \\&= \sfn{K-k}(r,:) \cdot
\left( 
	\begin{matrix} 
		\sfn{k}&\sfn{k}&\cdots&\sfn{k}
    \end{matrix} 
\right)v
\end{align*}
        where $r=\lfloor i/\delta^2\rfloor$ is the index of the row block of length $\delta^2$ from which row $i$ is drawn, $\sfn{K-k}(r,:)$ denotes the $r$th row of $S_{4^{K-k}}$, and the transform $\sfn{k}$ operates on a single $\delta\times \delta$ block of the image.   
        
        The measurement taken from the $i$th block can be written: 

$$b(i) = \sfn{K-k}(i,:) \cdot
\left( \begin{matrix} \sfn{k}(r',:) &\sfn{k}(r',:)&\cdots&\sfn{k}(r',:)
        \end{matrix} \right)v$$

where $r'= i-\delta^2r$ is the index of the obtained measurement relative to the start of the block. 

Now, note that $\sfn{k}(r'_i,:) v_k = \mu_k+e_k^{r'_i}.$  We then have

\begin{align*}b(i) &=\sfn{K-k}(i,:)  \mu +  \sfn{K-k}(i,:) 
\left( \begin{matrix} e_1^{r'_i} &e_2^{r'_i}&\cdots&e_{n^2}^{r'_i}
        \end{matrix} \right)
        \\&=\sfn{K-k}(i,:)  \mu+
        \eta_i
        \end{align*}
        where $ \eta_i=\sfn{K-k}(i,:)\left( \begin{matrix} e_1^{r'_i} &e_2^{r'_i}&\cdots&e_{n^2}^{r'_i}
        \end{matrix} \right).$       
        
        We have 
        $$Var(\eta_i) = \frac{1}{4^{K-k}}\sum_j Var(e_j^{r'_i}) = Mean_j(Var(e_j^{r'_i})).$$
        
The reconstructed preview is then
$$u_n(i) = \sfn{K-k} b = \mu +  \sfn{K-k} \eta.$$
Because $\sfn{K-k}$ is unitary and the entries in $\eta_i$ are identically distributed, each entry in $\sfn{K-k} \eta$ has the same variance as the entries of $\eta,$  which is  $Mean(Var(e_j^{r'_i})).$
  \end{IEEEproof}

\section{The 3DTV Model for Reconstruction}
\subsection{Motivation}
  First-generation compressive imaging exploits the compressibility of natural scenes in the spatial domain.  However, it is well-known that video is far more compressible than 2-dimensional images.  By exploiting this high level of compressibility, we can sample moving scenes at low rates without compromising reconstruction accuracy.

  Rather than attempt to exploit the precise pixel-to-pixel matching between images, we propose a model that allows adjacent images to share information without the need for an exact mapping (such as that obtained by optical flow).  The model, which we call 3DTV, assumes not only that images have small total variation in the spatial domain but also that each pixel generates an intensity curve that is sparse in time.

\begin{figure}
  \centering
  \includegraphics[width=7cm, height=7cm]{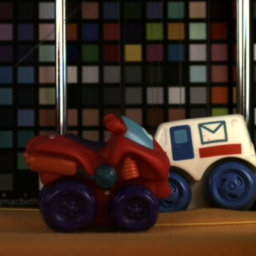}\\ \vspace{.1cm}
  \includegraphics[width=7cm, height=7cm]{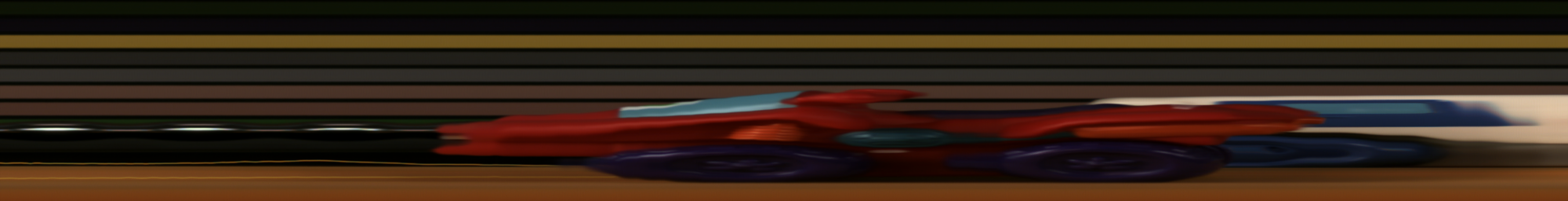}
  \caption{Piecewise constant frames form videos that are piecewise constant in time. (top)  A frame from our test video.  (bottom) We choose a column of pixels from the center of our test video, and plot this column over time.  The vertical axis is the ``y'' dimension, and the horizontal axis is time.  Note that the stationary background pixels are constant in time, while moving objects (such as the trucks) look similar in both the space and time domain resulting in a video that has small derivatives in the time direction.  }
   \label{fig:slice}
  \end{figure}

The 3DTV model is motivated by the following observation:  Videos with sparse gradient in the spatial domain also have sparse gradient in time.  TV-based image processing represents images with piecewise constant approximations.  Assuming piecewise constant image features, the intensity of a pixel only changes when it is crossed by a boundary.  As a result, applying TV in the spatial domain naturally leads to videos that have small TV in time.  This is demonstrated in Figure \ref{fig:slice}.

  The 3DTV model has several other advantages.  First, stationary objects under stable illumination conditions produce constant pixel values, and hence are extremely sparse in the time domain.  More importantly, the 3DTV model enforces ``temporal consistency'' of video frames; stationary objects appear the same in adjacent frames, and the flickering/distortion associated with frame-by-frame reconstruction is eliminated.
  
  \subsection{Video Reconstruction From Measurement Streams}

In practice, multiplexing cameras acquire one compressive measurement at a time.  Because data is being generated continuously and the scene is moving continuously in time, there is no ``natural'' way to break the acquired data into separate images.  For this reason, it makes sense to interpret the data as a ``stream'' -- an infinite sequence of compressive measurements denoted by $\{d_i\}_{i=1}^\infty$.  To reconstruct this stream, it must be artificially broken into sets, each of which forms the measurement data for a frame.   Such a decomposition can be represented graphically as follows:
$$\underbrace{d_1,d_2,d_3,d_4}_{b^1},\underbrace{d_5,d_6,d_7,d_8}_{b^2},\underbrace{d_9,d_{10},d_{11},d_{12}}_{b^3}\ldots$$
where $b^f$ denotes the measurement data used to reconstruct the $f$th frame.  In practice the data windows for each frame may overlap, or be different widths.  

Suppose we have collected enough data to reconstruct $F$ frames, denoted $\{u^f\}_{f=1}^F$.  The frames can then be simultaneously reconstructed using a variational problem of the form (\ref{energy}) with

  \eqb{blocks}
u = 
\left(
\begin{matrix}
u^1\\u^2\\
\vdots\\u^F
\end{matrix}
\right), \qquad
\Phi = 
 \left( 
 \begin{matrix}
  \Phi^1	 & 0 & \cdots  & 0\\
  0  & \Phi^2	&   \cdots  & 0\\
   \vdots&  & \ddots & \vdots \\
 0  & 0 &  \cdots  & \Phi^F
   \end{matrix}
   \right) \qquad
\eqe
 where $\Phi^f$ represents the measurement operator for each individual frame.

  \subsection{Mathematical Formulation}
  
  The 3DTV operator measures the Total-Variation of the video in both the space and time dimension.   
  The conventional Total-Variation semi-norm of a single frame $u^f$ is the absolute sum of the gradients:
    $$|\nabla u^f| =  \sum_{i,j} \sqrt{ (u^f_{i+1,j}-u^f_{i,j})^2+(u^f_{i,j+1}-u^f_{i,j})^2 }.$$
  The 3DTV operator generalizes this to the space and time dimensions
   \begin{align}\begin{split}
   \label{3dtv}|\nabla_3 u| &=  \sum_{i,j,f} \sqrt{ (u^f_{i+1,j}-u^f_{i,j})^2+(u^f_{i,j+1}-u^f_{i,j})^2 }\\&+|u^{f+1}_{i,j}-u^f_{i,j}|.
   \end{split}\end{align}
 
  We can reconstruct an individual frame using the variational model
  \eqb{single} u^f = \argmin_u \min  | \nabla u| +  \frac{\mu}{2}\|R^f S u  - b^f \|^2 \eqe
  where $R^f$ and $b^f$ denote the rows selector and data for frame $u^f$. 
  
  The 3DTV model extends conventional TV-based compressed sensing using the operator \eqref{3dtv}.  This model model can be expressed in a form similar to (\ref{single}) by stacking the frames into a single vector as in (\ref{blocks}).    We define combined row-selector and \stone transforms for all frames using the notation
  \begin{align*}
  R &= 
 \left( 
 \begin{matrix}
  R^1	 & 0 & \cdots  & 0\\
  0  &  R^2	&   \cdots  & 0\\
   \vdots&  & \ddots & \vdots \\
 0  & 0 &  \cdots  &  R^F
   \end{matrix}
   \right) \\
S &= 
 \left( 
 \begin{matrix}
  S_{N^2}	 & 0 & \cdots  & 0\\
  0  &  S_{N^2}	&   \cdots  & 0\\
   \vdots&  & \ddots & \vdots \\
 0  & 0 &  \cdots  &  S_{N^2}
   \end{matrix}
   \right).
 \end{align*}
Using this notation, the 3DTV model can be expressed concisely as
 \eqb{sit} \min  | \nabla_3 u | +  \frac{\mu}{2}\|RS u  - b \|^2 .\eqe
Note that just as in the single-frame case, $S$ is an orthogonal matrix and $R$ is diagonal.

\section{Numerical Methods For Compressive Recovery}
In this section, we discuss efficient numerical methods for image recovery.  There are many splitting methods that are capable of solving (\ref{sit}), however not all splitting methods are capable of exploiting the unique mathematical structure of the problem.  In particular, some methods require the solution of large systems using conjugate gradient sub-steps, which can be inefficient.  We focus on the Primal Dual Hybrid Gradient (PDHG) methods.  Because the \stone transform is self-adjoint, every step of the PDHG scheme can be written explicitly, making this type of solver efficient.

\subsection{PDHG}
Primal-Dual Hybrid Gradients (PDHG) is a scheme for solving minimax problems of the form
$$\max_y \min_x f(x)+\la Ax,y \ra-g(y)$$
where $f,g$ are convex functions and $A\in R^{m,n}$ is a matrix.   The algorithm was first introduced in \cite{EZC09}, and later in \cite{CP10}.  Rigorous convergence results are presented in \cite{HY12}.  Practical implementations of the method are discussed in \cite{GEB13}.

  The scheme treats the terms $f$ and $g$ separately, which allows the individual structure of each term to be exploited.  
The PDHG method in its simplest form is listed in Algorithm \ref{alg:pdhg}.
\begin{algorithm}[H]
\caption{PDHG}
\label{alg:pdhg}
\begin{algorithmic}[1]
\Require $x_0\in R^{N},\, y_0 \in R^m, \,  \tau>0, \sigma>0$
\For {$k=0,1, \ldots $}
\State $\hat x_{k+1} =x_k - \tau A^Ty_k$
\State $x_{k+1} = \argmin_x f(x)+\frac{1}{2\tau}\| x - \hat x_{k+1} \|^2$
\State $\bar x_{k+1} = x_{k+1}+(x_{k+1} - x_k)$
\State $\hat y_{k+1} =y_k + \sigma A \bar x_{k+1}$
\State $y_{k+1} = \argmax_y -g(y)-\frac{1}{2\sigma}\| y - \hat y_{k+1} \|^2$
\EndFor
\end{algorithmic}
\end{algorithm}
Algorithm \ref{alg:pdhg} can be interpreted as alternately minimizing for $x$ and then maximizing for $y$ using a forward-backward technique.  These minimization/maximization steps are controlled by two stepsize parameters, $\tau$ and $\sigma.$  The method converges as long as the stepsizes satisfy $\tau\sigma<\|A^TA\|.$ However the choice of $\tau$ and $\sigma$ greatly effects the convergence rate.  For this reason, we use the adaptive variant of PDHG presented in \cite{GEB13} which automatically tunes these parameters to optimize convergence for each problem instance.

\subsection{PDHG for Compressive Video}

In this section, we will customize PDHG to solve (\ref{sit}).  We begin by noting that 
$$|\nabla u| = \max_{p\in C}p\cdot \nabla u,$$
where $C=\{p| -1\le p_i\le 1\}$ denotes the $\ell_\infty$ unit ball.  Using this principle, we can write (\ref{sit}) as the saddle-point problem
\eqb{saddle}
\max_p \min_u  p\cdot \nabla u + \frac{\mu}{2}\| RHDu - s  \|^2 - 1_C(p)
\eqe
where $1_C(p)$ denotes the characteristic function of the set $C$, which is infinite for values of $p$ outside of $C$ and zero otherwise.

\begin{algorithm}[H]
\caption{PDHG Compressive Reconstruction}
\label{alg:recon}
\begin{algorithmic}[1]
\Require $u_0\in R^{N^2\times F},\, y_0 \in R^{3N^2\times F}, \,  \tau>0, \sigma>0$
\For {$k=0,1, \ldots $}
\State $\hat u_{k+1} =u_k - \tau \nabla^Ty_k$
\State $u_{k+1} = \argmin_u \frac{\mu}{2}\| RSu - b  \|^2+\frac{1}{2\tau}\| u - \hat u_{k+1} \|^2$
\State $\bar u_{k+1} = u_{k+1}+(u_{k+1} - u_k)$
\State $\hat p_{k+1} =p_k + \sigma \nabla \bar u_{k+1}$
\State $p_{k+1} = \argmax_p -1_C(p)-\frac{1}{2\sigma}\| p - \hat p_{k+1} \|^2$
\EndFor
\end{algorithmic}
\end{algorithm}

Note that only steps 3 and 6 of Algorithm \ref{alg:recon} are implicit.  The advantage of the PDHG approach (as opposed to e.g. the Alternating Direction Method of Multipliers) is that all implicit steps have simple analytical solutions.  

Step 3 of Algorithm \ref{alg:recon} in simply the projection of $\hat p_{k+1}$ onto the $\ell_\infty$ unit ball.  This projection is given by
$$p_{k+1} = \max\{\min\{p,1\},-1\}$$
where $\max\{\cdot\}$ and $\min\{\cdot\}$ denote the element-wise minimum and maximum operators.

Step 3 of Algorithm \ref{alg:recon} is the quadratic minimization problem  
 $$u_{k+1} = \argmin_u \frac{\mu}{2}\| RSu - b  \|^2+\frac{1}{2\tau}\| u - \hat u_{k+1} \|^2.$$ 
 The optimality condition for this problem is 
 $$\mu S^TR^T(RSu\kp -b)   + \frac{1}{\tau}(u\kp - \hat u)= 0$$
 which simplifies to
 $$(\mu S^TR^TRS +\frac{1}{\tau}I )  u\kp = \mu S^TR^Tb+\frac{1}{\tau} \hat u_{k+1}.$$
 If we note that $S$ and $R$ are symmetric, $R^2=R,$ and we write $I=S^2$ (because $S$ is symmetric and orthogonal) we get
  $$ S(\mu R +\frac{1}{\tau}I )S  u\kp = \mu SRb+\frac{1}{\tau} \hat u_{k+1}.$$
Since  $(\mu R +\frac{1}{\tau}I )$ is an easily invertible diagonal matrix, we can now write the solution to the quadratic program explicitly as
\eqb{explicit} u\kp =S (\mu R +\frac{1}{\tau}I )^{-1}S(\mu SRb+\frac{1}{\tau} \hat u_{k+1}).\eqe
Note that (\ref{explicit}) can be evaluated using only 2 fast \stone transforms.

\begin{figure*}
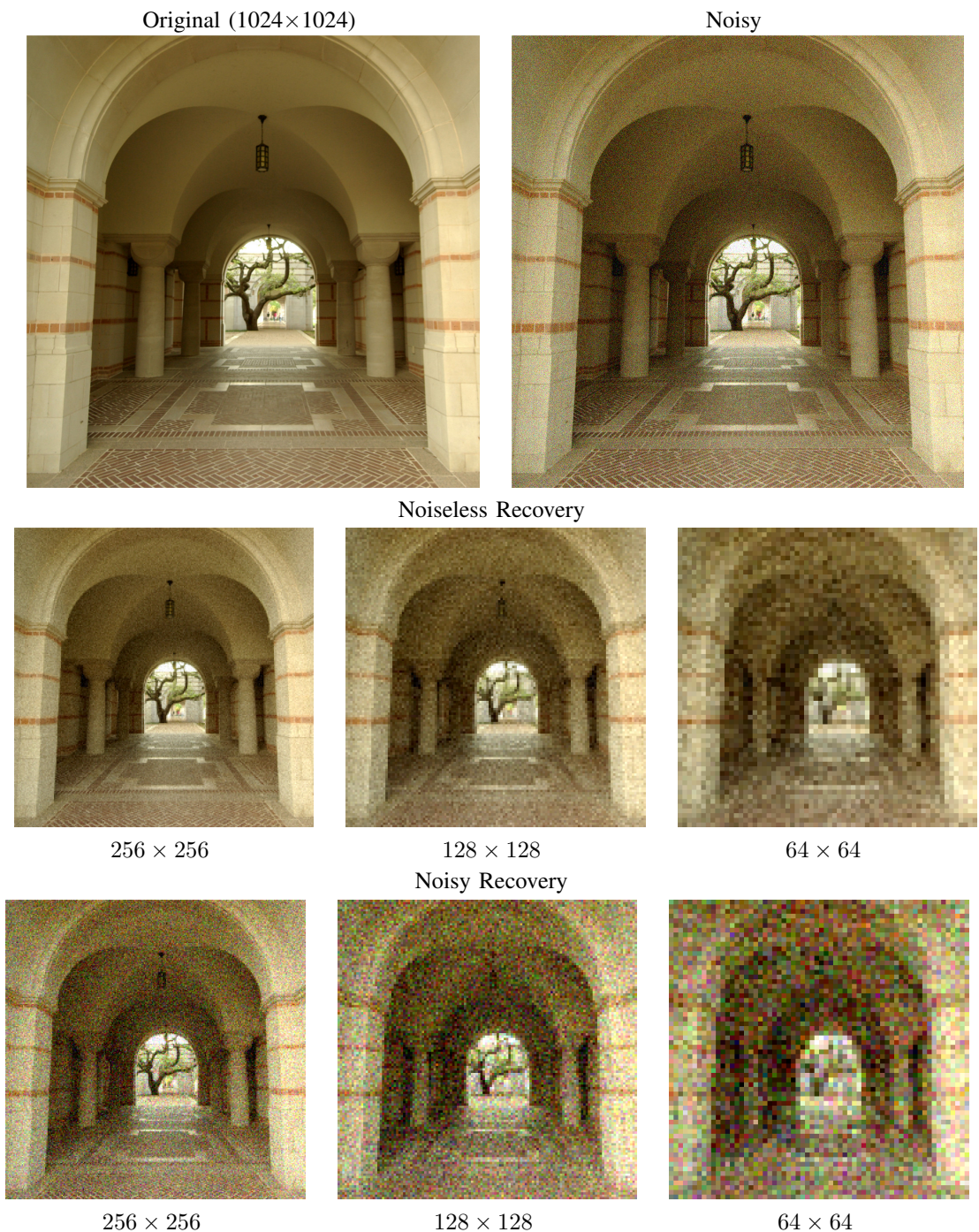
\centering
\labeledImage{arch_clean_1024}{ Original (1024$\times$1024)}{6.8cm}
\labeledImage{arch_dirty_1024}{ Noisy}{6.8cm}\\
\vspace{.2cm}Noiseless Recovery \vspace{.1cm}\\
\labeledImageBottom{arch_clean_256}{ $256\times 256$ }{4.5cm}
\labeledImageBottom{arch_clean_128}{ $128\times 128$  }{4.5cm}
\labeledImageBottom{arch_clean_64}{ $64\times 64$ }{4.5cm}\\
\vspace{.2cm} Noisy Recovery \vspace{.1cm}\\
\labeledImageBottom{arch_dirty_256}{ $256\times 256$  }{4.5cm}
\labeledImageBottom{arch_dirty_128}{ $128\times 128$ }{4.5cm}
\labeledImageBottom{arch_dirty_64}{ $64\times 64$ }{4.5cm}
\caption{Direct multiscale reconstruction from under-sampled transform coefficients.  (top left) The original image.  (top right) The image contaminated with Gaussian noise with standard deviation equal to $10\%$ of the peak image intensity. (center) Direct reconstruction of the noiseless measurements using under-sampled data.  (bottom) Direct reconstruction of the noisy data.   }
\label{fig:arch}
\end{figure*}

\section{Applications}
\subsection{Fast Object Recognition and the Smashed Filter}
  Numerous methods have been proposed to performing semantic image analysis in the compressive domain.   Various semantic tasks have been proposed including object recognition \cite{DDWLTKB07}, human pose detection \cite{KT12}, background subtraction \cite{CSDRB08} and activity recognition \cite{CXP10}.    These methods ascertain the content of images without a full compressive reconstruction.  Because these methods do not have access to an actual image representation, they frequently suffer from lower accuracy than classifiers applied directly to image data, and can themselves be computationally burdensome.
   
    When sensing is done in an MSS basis, compressive data can be quickly transformed into the image domain for semantic analysis rather than working in the compressive domain. This not only allows for high accuracy classifiers, but also can significantly reduce the computational burden of analysis.
      
        To demonstrate this, we will consider the case of object recognition using the ``Smashed Filter'' \cite{DDWLTKB07}.  This simple filter compares compressive measurements to a catalog of hypothesis images and attempts to find matches.  Let $\{H_i\}$ denote the set of hypothesis images, and $H_i(x)$ denote an image translated (shifted in the horizontal and vertical directions) by $x\in \reals^2.$  Suppose we observe a scene $s$ by obtaining compressive data of the form $b = \Phi s.$  The smashed filter analyzes a scene by evaluating
        \eqb{smash}\min_{x,H_i} \| \Phi s - \Phi H_i(x) \|^2 = \| b - \Phi H_i(x) \|^2.\eqe
   In plain words, the smashed filter compares the measurement vector against the transforms of every hypothesis image in every possible pose until a good $\ell_2$ match is found.  
   
   The primary drawback of this simple filter is computational cost.  The objective function \eqref{smash} must be explicitly evaluated for every pose of each test image.  The authors of  \cite{DDWLTKB07} suggest limiting this search to horizontal and vertical shifts of no more than 16 pixels in order to keep runtime within reason. 

\begin{figure*}
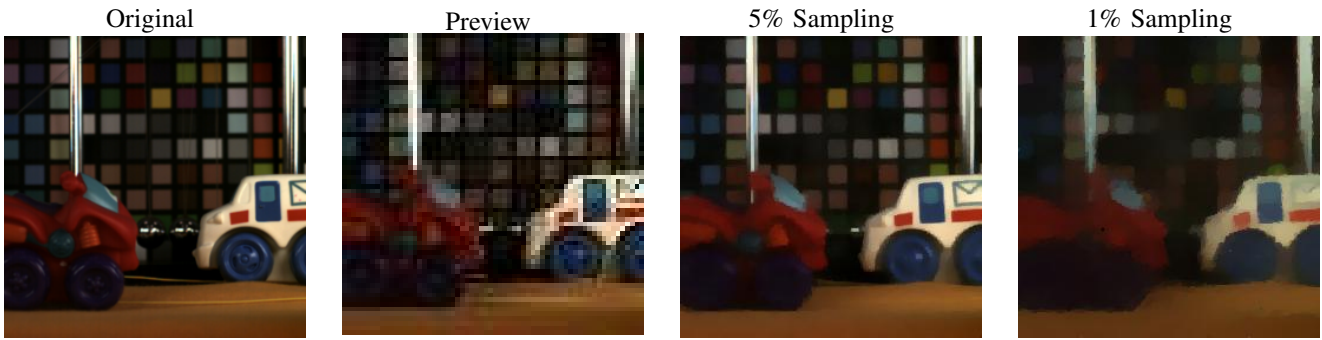

\centering
\labeledImage{original_2491}{Original}{4cm}
\labeledImage{64x64_28}{Preview}{4cm}
\labeledImage{compressed_05percent_image_58}{5\% Sampling }{4cm}
\labeledImage{compressed_01percent_image_58}{1\% Sampling }{4cm}
\caption{Reconstruction of high speed video from under-sampled data.   A frame from the original full resolution video is displayed on the left.  A $64\times 64$ preview generated from a stream with 6.25\% sampling is also shown.  Compressive reconstructions using 1\% and  6.25\% sampling are shown on the right. }
\label{fig:truckRecon}
\end{figure*}

The complexity and power of such a search can be dramatically improved using MSS measurements.  Suppose we obtain measurements of the form $b = RS_N s.$  We can then re-bin the measurements using Theorem \ref{thrm:recon}, and write them as $\bar b = S_n \bar s,$ where $\bar b$ is a dense vector of measurements and $\bar s$ is a lower resolution preview of the scene.  Applying Theorem \ref{thrm:recon} to the matching function \eqref{smash} yields 
 \begin{align}\begin{split}\label{smash2}
 \min_{x,H_i} \| R S_N  s - S_N H_i(x) \|^2 &=  \| S_n  \bar s - S_n \bar H_i(x) \|^2\\
 & =  \| \bar s -  \bar H_i(x) \|^2
 \end{split}\end{align}
where $\bar H_i$ is representation of image $H_i$ at resolution $n$.  The expression on the right of \eqref{smash2} shows that applying the smashed filter in the compressive domain is equivalent to performing classification in the image domain using the preview $\bar s$!  Furthermore, the values of $\| S_n  \bar s - S_n \bar H_i(x) \|^2 $ for all possible translations can be computed in $O(N\log N)$ using a Fourier transform convolution algorithm, which is a substantial complexity reduction from the original method. 
 
\subsection{Enhanced Video Reconstruction Schemes}
   Several authors propose compressive video reconstruction schemes that benefit from preview reconstruction. The basic concept of these methods is to use an initial low-quality reconstruction to obtain motion estimates.  These motion estimates are then used to build sparsity models that act in the time domain and are used for final video reconstruction.  Such methods include the results of Wakin \cite{PW13} and Sankaranarayanan \cite{SSB12} which rely on optical flow mapping as well as the iterative method of Asif \cite{AHBR13}.
   
      The methods proposed in \cite{PW13} and \cite{SSB12} use a Dual-Scale sensing basis that allows for the reconstruction of previews.  However, unlike the MSS framework proposed here, these matrices do not admit a fast transform.  The matrices proposed in \cite{SSB12} for example require $O(N^2)$ operations to perform a complete transform.  The proposed MSS matrices open the door for a variety of preview-based video reconstruction methods using fast $O(N\log N)$ transforms.

\section{Numerical Experiments}

\subsection{Preview Generation}

  We demonstrate preview generation using a simple test image.  The original image dimensions are 1024$\times$1024.  An MSS measurement operator is generated using the methods described in Section \ref{section:sto}.  The image is embedded into a vector using the pixel ordering generated by Algorithm \ref{alg:pixOrder}.  Transform coefficients are sampled in a structured random order generated by Algorithm \ref{alg:modeOrder}.
  
    We reconstruct previews by under-sampling the \stone coefficients, re-binning the results into a complete set of low resolution coefficients, and reconstructing using a single low-resolution \stone transform.   Two cases are considered.  In the first case we have the original noise-free image.  In the second case, we add a white Gaussian noise with standard deviation equal to $10\%$ of the peak signal (SNR = 20 Db).  Reconstructions are performed at $256\times 256,$ $128\times 128$ and $64\times 64$ resolutions.    Results are shown in Figure \ref{fig:arch}.  
  
  This example demonstrates the flexibility of MSS sensing --- the preview resolution can be adapted to the number of  available measurements.  The $64\times 64$ reconstructions  are obtained using only $64^2$ out of $1024^2$ measurements (less than 0.4\% sampling).

\subsection{Simulated Video Reconstruction}

To test our new image reconstruction framework, we use a test video provided by Mitsubishi Electric Research laboratories (MERL).  The video was acquired using a high-speed camera operating at 200 frames per second at a resolution of 256$\times$256 pixels.  A measurement stream was generated by applying the \stone transform to each frame, and then selecting coefficients from this transform.  Coefficients were sampled from the video at a rate of 30 kilohertz, which is comparable to the operation rate of currently available DMD's.  The coefficients were selected in the order generated by Algorithm \ref{alg:modeOrder}, and pixels were mapped into a vector with nested dissection ordering.  Thus, previews are available at any power-of-two resolution, and can be computed using a simple inverse \stone transform.  At the same time, the data acquired are appropriate for compressive reconstruction.

The goal is to reconstruct 20 frames of video from under-sampled measurements.  We measure the sampling rate as the percentage of measured coefficients per frame.   
For example, at the $1\%$ sampling rate, the total number of samples is $200\times 256^2\times 0.01.$  Results of compressive reconstructions at two different sampling rates are shown in Figure \ref{fig:truckRecon}.


Note that non-compressive reconstructions  require a lot of data (one measurement per pixel), and are therefore subject to the motion aliasing problems described in  \cite{Wakin12}.  By using the compressive reconstruction, we obtain high-resolution images from a small amount of data, yielding much better temporal resolution and thus less aliasing.  By avoiding the motion aliasing predicted by the classical analysis, the compressive reconstruction ``beats'' the Nyquist sampling limit.  

\begin{figure*}
\centering
\hspace{.4cm}Original \hspace{2.3cm}Complete Transform \hspace{2.3cm}Preview \hspace{2.8cm}Compressive \\
\includegraphics[trim  = 0cm 19cm 4cm 0cm, width = \linewidth, height=4.6cm, clip]{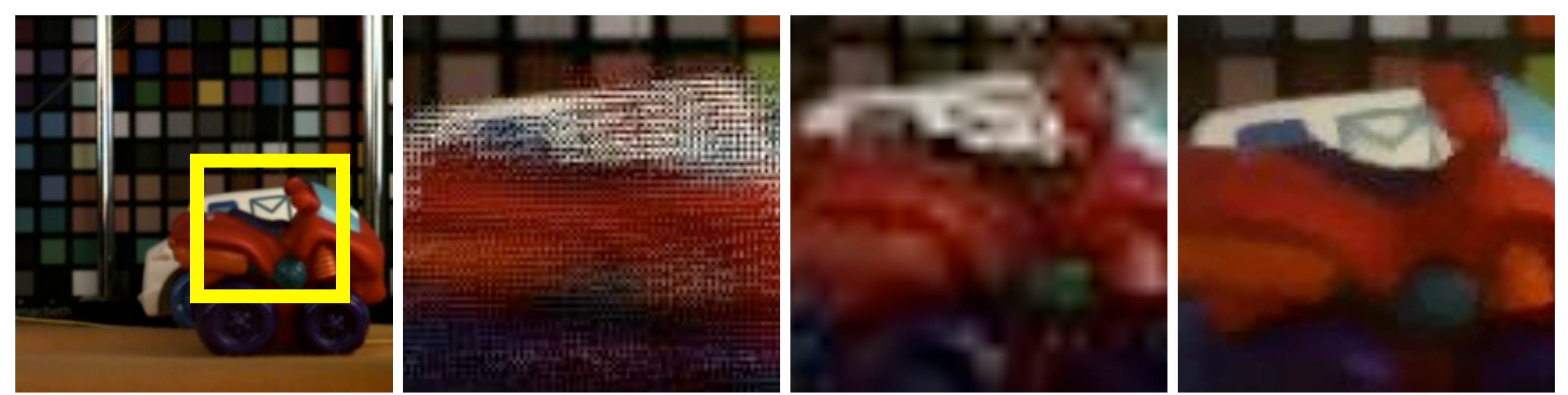}
\caption{Detailed views of reconstructions.   On the left is a frame from the original video showing the part of the image we will be using for comparisons. A full-resolution reconstruction using a complete set of transform coefficients is show beside it. Motion aliasing is present because of object motion over a long data acquisition time. A preview from 5\% sampling is also shown. Motion aliasing has been dramatically reduced by shortening the data acquisition time at the cost of lower resolution.  On the right is displayed a compressive reconstruction using the same 5\% sampling a the preview. This reconstruction simultaneously achieves high resolution and fast acquisition. }
\label{fig:detail}
\end{figure*}

Figure \ref{fig:detail} considers 3 different reconstructions at the same sampling rate.  First, a full-resolution reconstruction is created using a complete set of \stone samples.  Because a long measurement interval is needed to acquire this much data, motion aliasing is severe.  When a preview is constructed using a smaller number of samples and large pixels, motion aliasing is eliminated at the cost of resolution loss.  When the compressive/iterative reconstruction is used, the same under-sampled measurements reveal high-resolution details.

\subsection{Single-Pixel Video}
\begin{figure}
\centering
\begin{minipage}{8.5cm}
\centering
\begin{framed}
\hspace{-.3cm}
\includegraphics[width = 8cm]{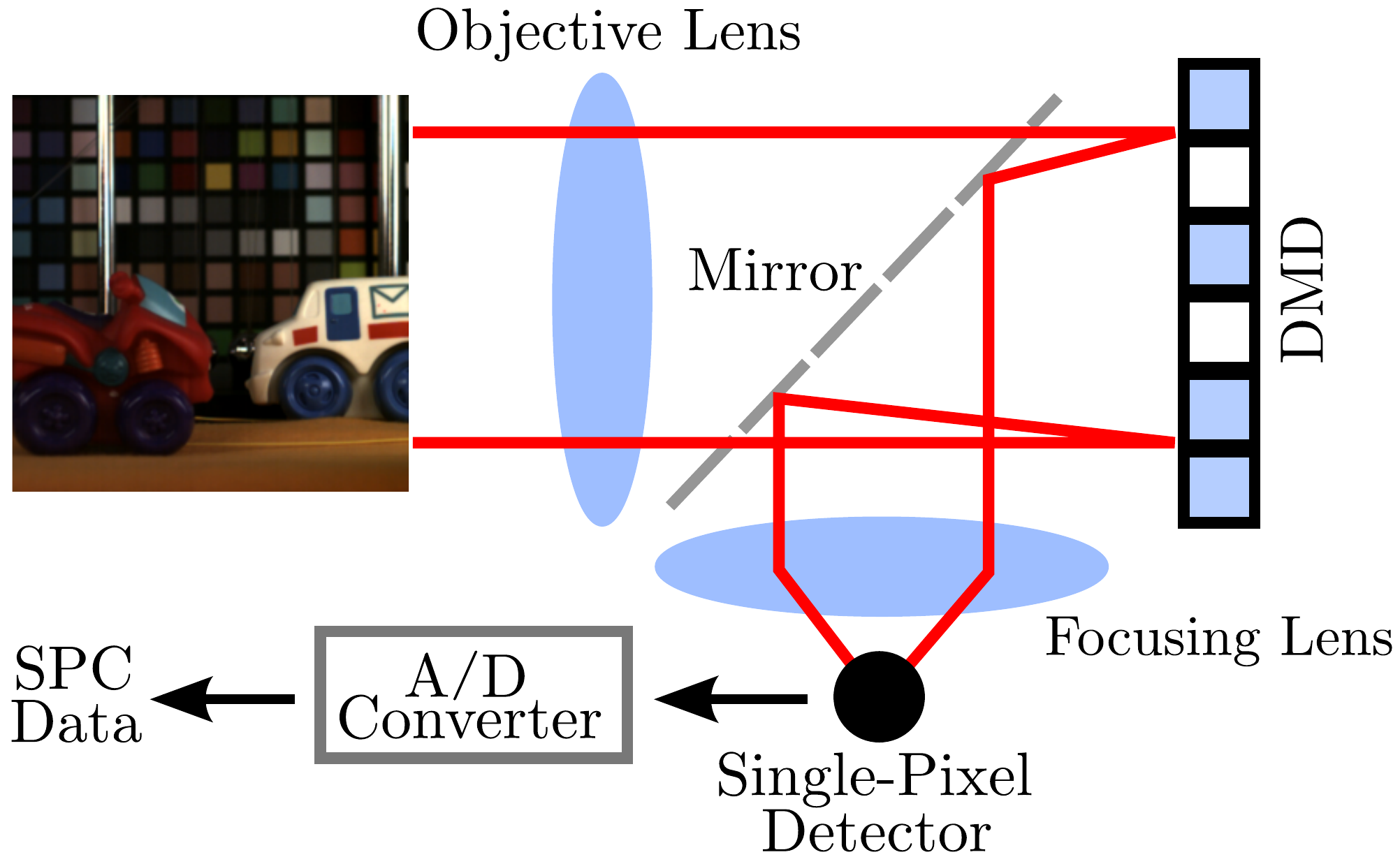}
\end{framed}
\end{minipage}
\begin{minipage}{8.5cm}
\centering
\begin{framed}
\vspace{-.2cm}
\hspace{-.40cm}
\includegraphics[width = 8.1cm, height=6cm, clip]{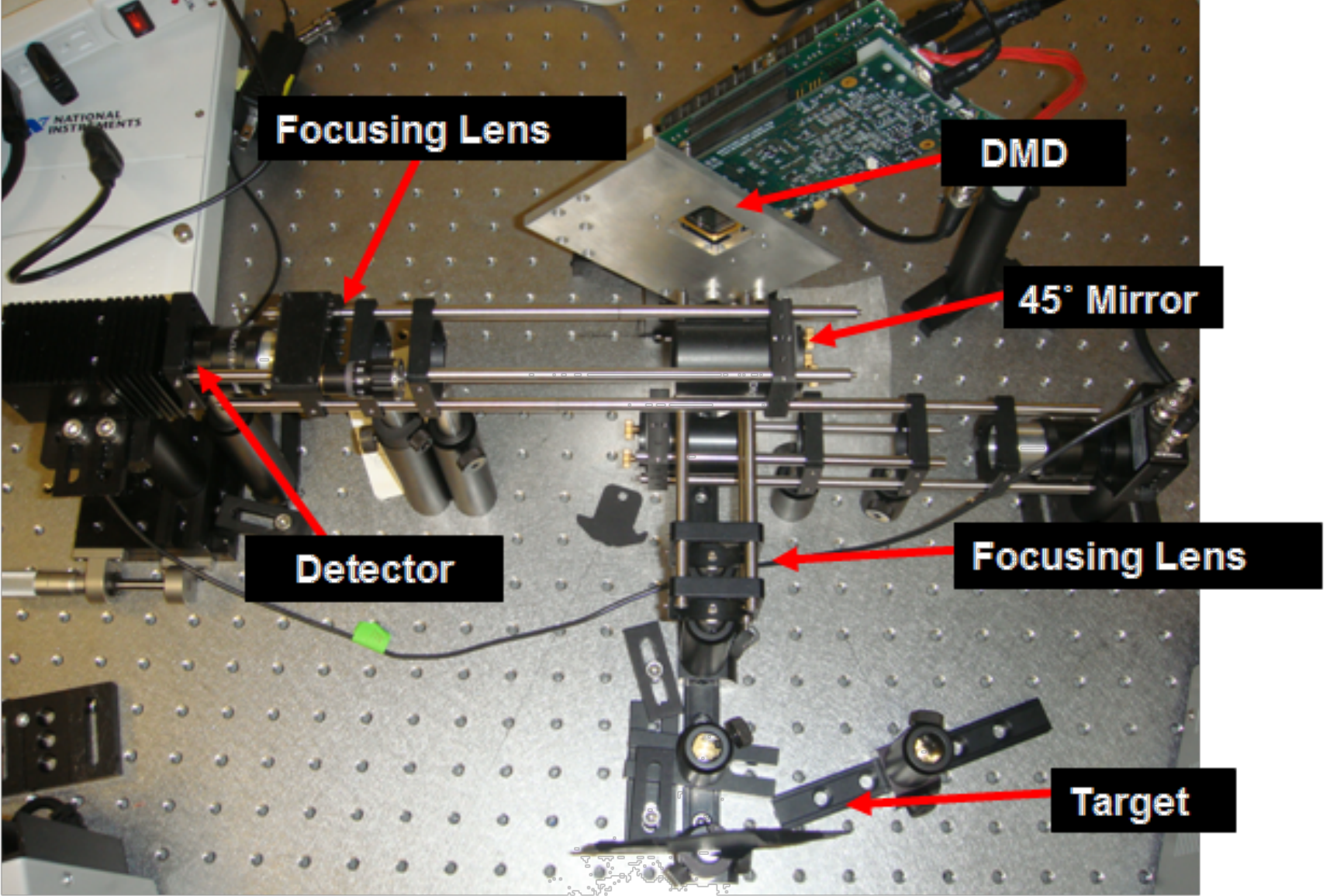}
\vspace{-.6cm}
\end{framed}
\end{minipage}
\caption{ Schematic drawing and table-top photo of a single-pixel camera.}
\label{fig:spc}
\end{figure}

  To demonstrate the \stone transform using real data, we obtained measurements using a laboratory setup.  The Rice single-pixel camera \cite{DDTLSKB08} is depicted in Figure \ref{fig:spc}.  The image of the target scene is focused by a lens onto a DMD. \stone transform patterns were loaded one-at-a-time onto the DMD in an order determined by Algorithm \ref{alg:modeOrder}. The DMD removed pixels with \stone coefficient $-1$ and reflects pixels with coefficient $+1$ towards a mirror.  A focusing lens then converges the selected pixels onto a single detector which generates a measurement. 
  
\begin{figure*}
\centering
\begin{minipage}{15cm}\raisebox{1.4cm}{\textbf{\huge A}}
\includegraphics[width = 3.5cm]{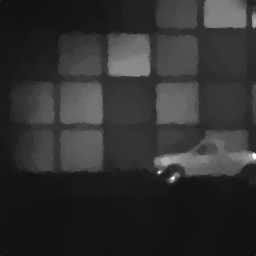}
\includegraphics[width = 3.5cm]{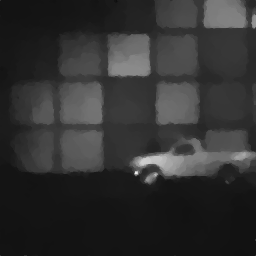}
\includegraphics[width = 3.5cm]{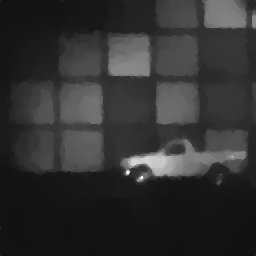}
\includegraphics[width = 3.5cm]{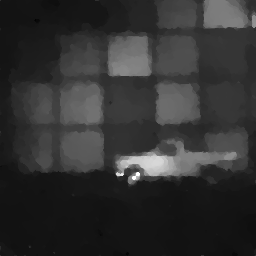}  
\end{minipage}\\
\vspace{.1cm}
\begin{minipage}{15cm}\raisebox{1.4cm}{\textbf{\huge B}}
\includegraphics[width = 3.5cm]{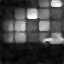}
\includegraphics[width = 3.5cm]{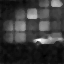}
\includegraphics[width = 3.5cm]{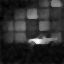}
\includegraphics[width = 3.5cm]{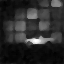}
\end{minipage}\\
\vspace{.1cm}
\begin{minipage}{15cm}\raisebox{1.4cm}{\textbf{\huge C}}
\includegraphics[width = 3.5cm]{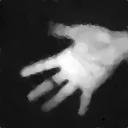}
\includegraphics[width = 3.5cm]{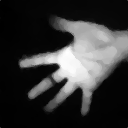}
\includegraphics[width = 3.5cm]{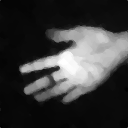}
\includegraphics[width = 3.5cm]{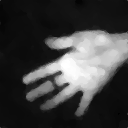}
\end{minipage}\\
\vspace{.1cm}
\begin{minipage}{15cm}\raisebox{1.4cm}{\textbf{\huge D}}
\includegraphics[width = 3.5cm]{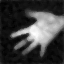}
\includegraphics[width = 3.5cm]{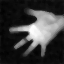}
\includegraphics[width = 3.5cm]{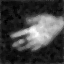}
\includegraphics[width = 3.5cm]{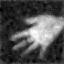}
\end{minipage}
\caption{Video reconstructions from the single-pixel camera. (A/C) High-resolution compressive reconstructions of the ``car'' and ``hand'' scenes. (B/D)  64$\times$64 previews. }
\label{fig:dmd}
\end{figure*}

  Data was generated from two scenes at two different resolutions.  The ``car'' scene was acquired using 256$\times$256 \stone patterns, and the ``hand'' scene was acquired using $128\times128$ patterns.  For both scenes, previews were reconstructed at $64\times 64$ resolution.  Full-resolution compressive reconstructions were also performed. Frames from the resulting reconstructions are displayed in Figure \ref{fig:dmd}.  Both the previews and compressive reconstructions were generated using the same measurements.  Note the higher degree of detail visible in the compressive reconstructions.  

\section{Conclusion}

Compressed sensing creates dramatic tradeoffs between reconstruction accuracy and reconstruction time.  While compressive schemes allow high-resolution reconstructions from under-sampled data, the computational burden of these methods prevents their use on portable embedded devices.  The \stone transform enables immediate reconstruction of compressive data at Nyquist rates.  The same data can then be ``enhanced''  using compressive schemes that leverage sparsity to ``beat'' the Nyquist limit.  
  
  The multi-resolution capabilities of the  \stone transform are paramount for video applications, where data resources are bound by time constraints.  The limited sampling rate of compressive devices leads to ``smearing'' and motion aliasing when sampling high-resolution images at Nyquist rates.  We are left with two options:  either slash resolution to decrease data requirements, or use compressive methods that prohibit real-time reconstruction. The \stone transform offers the best of both worlds:  immediate online reconstructions with high-resolution compressive enhancement. 

\section*{Acknowledgments}
The authors would like to thank Yun Li for his help in the lab, and Christoph Studer for many useful discussions.  This work was supported by the Intelligence Community (IC) Postdoctoral Research Fellowship Program.

\bibliography{/Users/Tom/Documents/latexDocs/bib/tom_bibdesk}

\end{document}